\documentclass[11pt,a4paper]{article}
 
\usepackage[a4paper,left=26mm,right=26mm,top=24mm,bottom=24mm]{geometry}
\usepackage[T1]{fontenc}
\usepackage{helvet}

\usepackage{amsmath,amssymb}
\usepackage{microtype}
\usepackage{xcolor}
\usepackage[super,comma,sort&compress]{natbib}
\usepackage{titlesec}
\usepackage{enumitem}
\usepackage{booktabs,tabularx,array}
\usepackage{caption}
\usepackage{tikz}
\usetikzlibrary{arrows.meta,positioning,fit,backgrounds}
\usepackage[most]{tcolorbox}
\usepackage[hidelinks]{hyperref}
 
\definecolor{natred}{HTML}{C8102E}
\definecolor{boxbg}{HTML}{EEF3F8}
\definecolor{boxline}{HTML}{1F4E79}
 
\titleformat{\section}{\large\bfseries\color{natred}}{}{0pt}{}
\titlespacing*{\section}{0pt}{1.4ex plus .4ex}{0.7ex}
\titleformat{\paragraph}[runin]{\bfseries}{}{0pt}{}[.\ ]
\titlespacing*{\paragraph}{\parindent}{0.8ex plus .2ex}{0.4em}
 
\DeclareCaptionLabelSeparator{bar}{\ $|$\ }
\newtcolorbox{naturebox}[1]{
  enhanced, breakable, colback=boxbg, colframe=boxline,
  boxrule=0pt, leftrule=2.2pt, arc=0pt, left=6pt, right=6pt, top=5pt, bottom=5pt,
  fontupper=\small, before skip=8pt, after skip=8pt,
  title={\textbf{#1}}, coltitle=boxline, colbacktitle=boxbg,
  attach boxed title to top left={yshift=0pt}, boxed title style={boxrule=0pt,colback=boxbg,left=0pt,top=0pt,bottom=2pt},
  fonttitle=\normalsize
}
 
\hypersetup{pdftitle={Deep reinforcement learning for fluid
flow control: challenges and future
directions}}

\begin{document}

{%
\begin{center}
\begin{minipage}{\textwidth}
\raggedright
{\footnotesize\bfseries\color{natred}PERSPECTIVE}\par\vspace{4pt}
{\fontsize{16}{22}\selectfont\bfseries Glass-Box Deep Reinforcement Learning for High-Speed Flow Control
 \par}\vspace{7pt}
{\Large Ameya D. Jagtap \\ 
\small Aerospace Engineering Department, Worcester Polytechnic Institute, Worcester, MA 01609, USA \\ Emails: ajagtap@wpi.edu, ameyadjagtap@gmail.com}\par
\vspace{6pt}
\rule{\textwidth}{0.6pt}\par\vspace{6pt}
\begin{minipage}{0.96\textwidth}
\small\noindent\textbf{Abstract:\ \ }%
\textbf{Deep reinforcement learning (DRL) is challenging long-held assumptions in active flow control. By learning directly from interaction with a flow, DRL agents can find actuation strategies for nonlinear, unsteady flows without explicit models, often exploiting mechanisms that classical control would not anticipate. High-speed flows are a demanding test of this promise: shocks, shock/boundary-layer interactions, transition, aerothermal loading and microsecond-scale dynamics strain both algorithms and hardware. Yet most demonstrations remain idealized, low-speed simulations with limited evidence of robustness, interpretability, transferability or experimental deployment. In this perspective, we examine what constrains DRL-based high-speed flow control, including compressible simulation cost and fidelity, data efficiency, sensing in extreme environments, reward design, explainability, and generalization across Mach and Reynolds number. We argue that these limits reflect a tension between data-driven optimization and the structured physics, constraints and uncertainties of real flows. We then highlight where DRL may go beyond conventional approaches: adaptive control of shock-induced separation and turbulent boundary layers, coordination of distributed actuators, and discovery of unconventional flow-manipulation mechanisms. \textbf{The central challenge, however, is translating these capabilities into reliable technology, which will require validated compressible simulation environments, physics- and constraint-aware learning, robustness to off-design conditions and sensor degradation, real-time operation, hardware-in-the-loop and wind-tunnel validation, and credible pathways to safety and certification.} This demands closer integration of fluid mechanics, control theory, machine learning and experiment, positioning DRL not as a replacement for classical control but as a complementary technology whose value will be set by its ability to operate reliably under zero-shot learning.}
\end{minipage}
\end{minipage}
\end{center}
\vspace{4pt}
}
\vspace{-0.5cm}
\begin{naturebox}{Key points}
\begin{itemize}[leftmargin=1.1em,itemsep=2pt,topsep=2pt]
\item DRL can discover effective flow-control strategies through interaction, but its performance is bounded by the fidelity of the environment in which it learns.
 
\item High-speed flows, with shocks, shock--boundary-layer interaction, transition and thermal loads, make training environments costly and sensitive to numerical choices.
 
\item Numerical artefacts can become exploitable features of the control problem, making simulation fidelity a prerequisite for credible technology readiness.
 
\item Model-based learning is data-efficient in simple flows, but its advantage can diminish in turbulent, high-dimensional and shock-dominated regimes where model errors accumulate.
 
\item Generalization across Mach and Reynolds numbers, geometries and sensing conditions, together with robustness to noise, latency and actuator limits, remains a major barrier to deployment.
 
\item Explainable and interpretable policies are needed to build trust, reveal the flow physics being exploited and support certification.
 
\item Physics-aware, hybrid and foundation-model approaches could provide more generalizable and constraint-aware controllers.
 
\item Higher technology readiness requires real-time operation, hardware compatibility, experimental validation and safety beyond nominal simulation conditions, especially at high speed.
\end{itemize}
\end{naturebox}

\section{Introduction}
Controlling a fluid flow is among the oldest ambitions of fluid mechanics, and among the most consequential for engineering. Delaying separation on a wing, suppressing vortex shedding behind a bluff body or damping unsteady shock motion can cut drag, noise and structural loads. Yet the very properties that make flows worth controlling also make them hard to control: strong nonlinearity, a wide range of coupled spatial and temporal scales, sensitivity to small perturbations and, in practice, access to only a few noisy measurements. Classical active flow control (AFC) responds with open-loop forcing tuned empirically, or with feedback laws derived from linearized or reduced-order models\cite{brunton2015closed,brunton2020data}. These approaches work well when the dynamics are dominated by a handful of coherent structures around a known base state, and degrade away from it.

Deep reinforcement learning (DRL) offers a different bargain. Instead of modelling the flow and then designing a controller, a DRL agent learns a control policy by trial and error, guided only by a reward that encodes the objective (Box~1). Early studies showed that such agents can suppress vortex shedding behind a cylinder or outperform hand-designed strategies\cite{rabault2019artificial,fan2020reinforcement}, and the approach has since been extended to a growing range of configurations\cite{wan2021rlcontrol,xia2024active}. The appeal is clear: no explicit model is required, the objective can be non-differentiable or only sparsely available, and the learned law is free to exploit nonlinear mechanisms that linear analysis cannot reach.

The same appeal extends to high-speed flows, where shock waves, thermodynamic effects and fluid--structure interaction (FSI) make accurate reduced-order models especially difficult to build. In the transonic regime, DRL agents have been used to suppress buffet and buffeting in coupled FSI systems without an explicit dynamic model\cite{ren2024adaptive}, to control airfoil flutter with synthetic jets and extend the flutter boundary beyond the training conditions\cite{gong2024active}, and to reduce drag on an RAE2822 airfoil, with and without lift penalties, using a high-order discontinuous Galerkin solver as the learning environment\cite{mondal2025shocks}. These efforts contrast with model-based strategies such as closed-loop shock-control bumps driven by lift feedback\cite{deng2023closed} and adaptive control built on time-variant reduced-order models\cite{gao2025transonic}, which depend on the quality of the underlying model. In the hypersonic regime, the emphasis shifts from aerodynamic efficiency to propulsion stability. Here DRL has been applied to microjet control of shock-wave/laminar-boundary-layer interaction on a compression ramp, using a recurrent surrogate of the flow to avoid the prohibitive cost of hypersonic CFD during training\cite{tao2025control}. Higher-order schemes are employed to control hypersonic unstart within the isolator under uncertainty in back pressure \cite{mondal2026hypersonic}. It has also been combined with active disturbance rejection control to regulate scramjet fuel pulse injection, meeting thrust demands while preventing unstart\cite{liang2025research}, with the agent trained by the twin delayed deep deterministic policy gradient (TD3) algorithm\cite{fujimoto2018addressing}. Closely related work on sensing shows how limited the available information can be: deep learning can reconstruct near-wall pressure fluctuations in shock--boundary layer interaction from sparse, under-sampled data\cite{poulinakis2023deep}, and unstart in combined-cycle inlets can be predicted from only two to four surface pressure sensors\cite{min2024machine}. High-speed flows are therefore a demanding test bed for DRL: simulations are costly, sensing is sparse and failure, such as unstart or flutter, is expensive.

Most reported successes in DRL-based AFC, however, come from low-Reynolds-number, two-dimensional simulations with idealized sensing and actuation, and even the high-speed studies above are largely numerical. It is far less clear whether the resulting policies are robust, whether they transfer to new conditions, and whether they can be trusted, let alone understood, in a wind tunnel or in flight. In this perspective, we examine these questions in turn. We argue that these challenges are manifestations of the broader \textit{sim-to-real gap}: policies that perform well in controlled numerical environments may fail when confronted with the complexity, variability, and uncertainty of real fluid systems. The gap is especially wide for compressible and hypersonic flows, where high-fidelity training environments are expensive, surrogate models introduce their own errors and experimental validation is difficult. Closing this gap will require more than improved learning algorithms; it will require methods that account for the differences between simulated and physical environments and can maintain reliable performance under those differences. How effectively this sim-to-real gap is addressed will determine whether DRL can mature into a practical flow-control paradigm or remain largely a computational curiosity.

\begin{figure}[t]
\centering
\resizebox{\linewidth}{!}{%
\begin{tikzpicture}[
  font=\footnotesize\sffamily,
  box/.style={draw=boxline, fill=boxbg, rounded corners=2pt, thick, align=center, minimum height=11mm, inner sep=5pt},
  arr/.style={-{Stealth[length=2.2mm]}, thick, color=boxline},
  lbl/.style={font=\scriptsize\itshape, color=black!75}
]
\node[box, minimum width=42mm, minimum height=24mm] (agent) at (0,0)
  {\textbf{DRL agent}\\[1pt]neural-network policy\\$\pi_\theta(a_t\mid s_t)$};
\node[box, minimum width=52mm] (sim) at (7.6,1.9)
  {\textbf{Simulated environment}\\CFD solver (URANS, LES, DNS)};
\node[box, minimum width=52mm, fill=natred!8, draw=natred] (exp) at (7.6,-1.9)
  {\textbf{Physical environment}\\wind tunnel or flight article};

\draw[arr] (2.1,0.8) -- node[above,sloped,lbl]{action $a_t$} (5.0,2.2);
\draw[arr] (5.0,1.6) -- node[below,sloped,lbl]{state $s_t$, reward $r_t$} (2.1,0.3);
\draw[arr, color=natred] (2.1,-0.3) -- node[above,sloped,lbl]{action $a_t$} (5.0,-1.6);
\draw[arr, color=natred, dashed] (5.0,-2.2) -- node[below,sloped,lbl]{noisy $s_t$, delay} (2.1,-0.8);
\draw[<->, thick, dashed, color=natred] (7.6,1.2) -- (7.6,-1.2);
\node[anchor=west, font=\scriptsize, color=natred, align=left] at (7.75,0)
  {\textbf{simulation-to-reality gap:}\\[-1pt]idealized boundary conditions, full observability\\[-1pt]and numerical artefacts vs.\ noise, latency, actuator limits};
\end{tikzpicture}}
\caption{\textbf{The DRL control loop and its environment dependence}: The agent observes sensor measurements and a reward from its environment and returns an actuation command. In simulation (top), the CFD solver defines the dynamics, the observable state and the reward, so any numerical artefact it contains can be learned and exploited. In a physical environment (bottom), the same policy meets noise, latency and actuator limits that were absent in training. This \textit{simulation-to-reality gap} is the central obstacle to deployment.}
\label{fig:loop}
\end{figure}
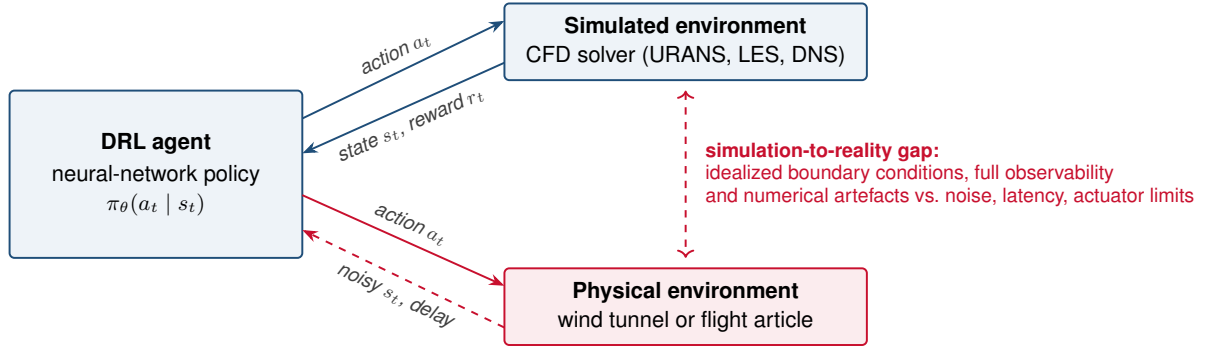

\begin{naturebox}{Box 1 $|$ Deep reinforcement learning for flow control in brief}
In DRL, an agent observes the state of an environment through a set of measurements $s_t$ (for example wall-pressure or shear-stress probes), applies an action $a_t$ (jet blowing, surface motion or plasma actuation) and receives a scalar reward $r_t$ (for example reduced drag penalized by actuation power). A neural-network policy $\pi_\theta(a_t\mid s_t)$ is trained to maximize the expected discounted return
\begin{equation*}
J(\theta)=\mathbb{E}_{\pi_\theta}\Big[\sum_{t=0}^{\infty}\gamma^{t}r_t\Big],
\end{equation*}
where $\gamma\in[0,1)$ is a discount factor. In AFC the environment is usually a CFD solver and, more rarely, an experiment (Fig.~\ref{fig:loop}).\vspace{3pt}

\textbf{Model-free} methods learn the policy directly from interaction data. \textbf{Model-based} methods additionally learn, or are given, a model of the transition dynamics and use it for planning or for generating synthetic experience. \textbf{On-policy} algorithms learn only from data generated by the current policy, whereas \textbf{off-policy} algorithms reuse past experience, which usually improves sample efficiency.
\end{naturebox}

\section{The simulator is the environment}
The performance and reliability of DRL for AFC in high-speed flows are set, to a remarkable degree, by the compressible solver used for training. The solver defines the state space, governs the transition dynamics and determines the reward. Any systematic numerical or modelling error is therefore embedded in the learned policy. Unlike conventional optimization, DRL can amplify such errors: if a consistent numerical artefact yields an apparent gain in reward, the agent will find and exploit it \cite{amodei2016concrete}, producing control strategies that are non-physical or that fail outside the training environment.

\begin{figure}[htpb]
\centering
\includegraphics[trim=1cm 0.5cm 1cm 0cm, scale=0.35, clip=true]{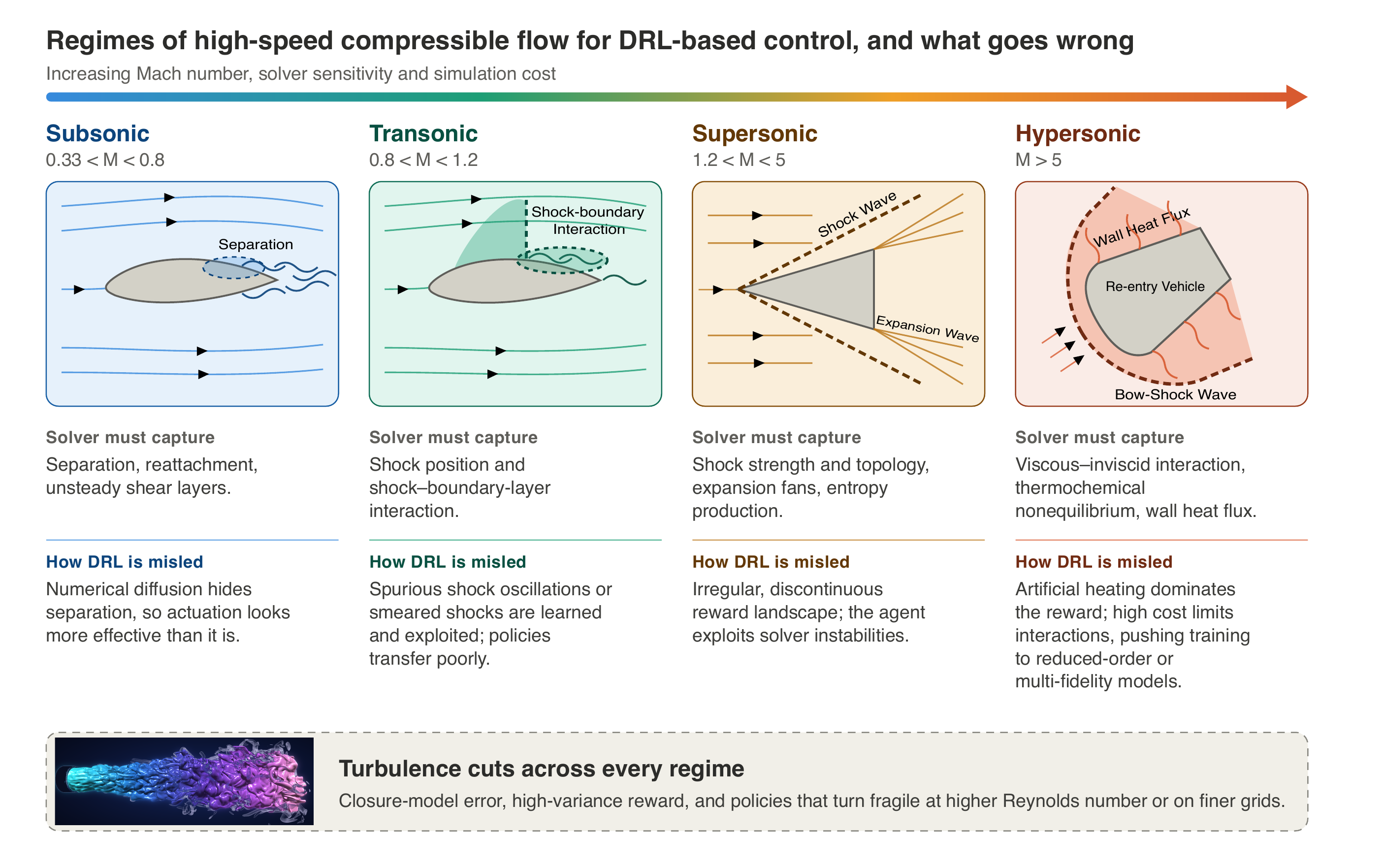}
\caption{\textbf{Regimes of high-speed compressible flow for DRL-based control}: Solver sensitivity and simulation cost increase with Mach number, from separation and turbulence-closure errors (subsonic), to buffet and low-frequency shock motion (transonic), to shock--boundary-layer interaction and closure-dependent artefacts (supersonic), to transition, thermal loads and prohibitive scale-resolving cost (hypersonic).}
\label{fig:F1}
\end{figure}

In high-speed flows this coupling is especially strong, because the quantities that matter for control are precisely those that compressible solvers capture least reliably. Shock-capturing schemes rely on upwinding, limiters or nonlinear reconstruction, which introduce dissipation that is both solution-dependent and nonlinear \cite{pirozzoli2011numerical,jiang1996efficient}. Shock position, shock/boundary-layer interaction (SBLI) unsteadiness and separation-bubble dynamics are sensitive to the flux scheme, the limiter and the turbulence closure \cite{dolling2001fifty,gaitonde2015progress}. At hypersonic speeds, wall heat flux and boundary-layer transition depend further on grid alignment, thermochemical modeling and transition prediction \cite{peter2001computational,Fedorov2011}. Because DRL optimizes cumulative reward over long horizons, small errors in the predicted response to actuation accumulate over many control steps, and the agent may end up stabilizing numerical features, such as a limiter-switching pattern or a smeared shock, rather than true flow structures. Solver consistency and physical fidelity are therefore more critical for DRL than for most closed-loop or adjoint-based methods. Moreover, the cost of high-fidelity high-speed simulation (scale-resolving turbulence, finite-rate chemistry) limits the number of environment interactions available for training \cite{slotnick18332cfd}. The consequences differ across speed regimes (Table~\ref{tab:regimes} and Figure ~\ref{fig:F1}).

\begin{table*}[htpb]
\centering
\small
\renewcommand{\arraystretch}{1.25}
\caption{\textbf{Solver requirements and failure modes of DRL control in high-speed flow regimes}}
\label{tab:regimes}
\begin{tabularx}{\textwidth}{@{}>{\raggedright\bfseries}p{22mm} >{\raggedright\arraybackslash}X >{\raggedright\arraybackslash}X >{\raggedright\arraybackslash}p{38mm}@{}}
\toprule
Regime & What the solver must capture & How errors can mislead the agent & Key references \\
\midrule
Turbulence and transition (all regimes)
& Compressible near-wall shear, coherent structures, separation; closure choice (RANS, URANS, hybrid RANS--LES, wall-modelled LES); transition onset in supersonic and hypersonic boundary layers
& Agent controls closure-model artefacts; excess dissipation suppresses fluctuations and inflates apparent control authority; RANS failure in separated and shock-interacting flow; policies fragile at higher Reynolds number or finer grids
& \cite{spalart2000strategies,duraisamy2019turbulence,larsson2016large,bodart2011wall,garnier2009large} \\
Subsonic / low Mach
& Separation, reattachment, unsteady shear layers and vortex shedding; correct low-Mach behaviour of compressible solvers
& Compressible schemes at low Mach number are over-dissipative unless preconditioned or low-dissipation; numerical diffusion damps shedding, so actuation effectiveness is overestimated; 2D-trained policies may exploit coherent shedding absent in 3D
& \cite{turkel1999preconditioning,wiliamson1996vortex,rabault2019artificial,vignon2023recent} \\
Transonic
& Shock--boundary-layer interaction, shock position and motion, buffet onset and low-frequency shock oscillation
& Buffet onset and shock location depend strongly on closure and scheme; spurious shock oscillations or smeared shocks are learned and exploited; poor transfer to higher fidelity or experiment
& \cite{lee2001self,crouch2009origin,dolling2001fifty,clemens2014low} \\
Supersonic
& Shock strength and topology, shock/shock and SBLI separation bubbles, expansion fans, entropy production
& Limiter switching, Gibbs-type oscillations and carbuncle-type instabilities create an irregular, discontinuous reward landscape; agent exploits solver instabilities; SBLI response to actuators (e.g.\ vortex generators) is closure-sensitive
& \cite{quirk1994contribution,pandolfi2001numerical,jiang1996efficient,gaitonde2015progress,souverein2010effect} \\
Hypersonic
& Viscous--inviscid interaction, high-temperature real-gas effects, thermochemical nonequilibrium, wall heat flux, boundary-layer transition and its control (e.g.\ porous coatings)
& Artefacts such as artificial heating (flux-scheme- and grid-alignment-dependent) dominate the reward; transition-location errors mislead stability-based control; high cost limits interactions, favouring reduced-order or multi-fidelity training
& \cite{park1989nonequilibrium,anderson1989hypersonic,candler1991computation,kitamura2013towards,Fedorov2011,bertin2003fifty,peherstorfer2016survey,mondal2026hypersonic} \\
\bottomrule
\end{tabularx}
\end{table*}

DRL can discover effective flow-control strategies through interaction, but its performance is bounded by the fidelity of the environment in which it learns. Two solver properties matter most: what the solver lets the agent observe, set by the turbulence closure (URANS, hybrid RANS--LES or wall-modeled LES), and how much it lets the agent explore, limited by computational cost. Errors in near-wall modeling propagate directly into wall shear stress, pressure fluctuations and shock motion, the very quantities that most control objectives target \cite{spalart2000strategies,larsson2016large,bodart2011wall,duraisamy2019turbulence}. High-order, low-dissipation numerics and adequate temporal resolution are essential, because excess dissipation suppresses fluctuations, inflates apparent control authority and yields policies that are fragile at higher Reynolds number, on refined grids or in experiment. High-speed flows, with shocks, shock--boundary-layer interaction, low-frequency shock motion, transition and thermal loads, raise reward variance and make training costly and numerically sensitive \cite{clemens2014low}, and numerical artefacts can become exploitable features of the control problem. Scale-resolving and thermochemically complete simulations restrict the number of environment interactions, pushing practitioners towards reduced-order surrogates and multi-fidelity training \cite{slotnick18332cfd,peherstorfer2016survey}. These improve data efficiency, but their advantage can diminish in turbulent, shock-dominated regimes where model errors accumulate. Generalization across Mach and Reynolds numbers, geometries and sensing conditions, together with robustness to noise, latency and actuator limits, remains a major barrier to deployment, and explainable, interpretable policies are needed to build trust and support certification. Physics-aware, hybrid and foundation-model approaches \cite{menon2026scientific}, along with machine-learning-accelerated solvers \cite{kochkov2021machine}, could provide more generalizable controllers, but remain largely untested for shock-dominated flows. High-fidelity simulation can act as a virtual laboratory for assessing actuators, sensing and control frequency, yet higher technology readiness ultimately requires real-time operation, hardware compatibility, experimental validation and safety beyond nominal simulation conditions.

\section{Model-free and model-based learning}
\vspace{0.2cm}
\noindent \textbf{Where model-based DRL works}:
Model-based DRL promises better data efficiency and a degree of interpretability by learning, or exploiting, a model of the flow and optimizing actions over an extended horizon. In AFC it has been implemented with learned surrogates, reduced-order representations and, in some cases, numerical PDE models\cite{li2021learning,halder2020deep,dong2023surrogate,portal2023active,ye2024reducing,mao2024model,zhang2024efficient,wang2024dynamic,werner2024numerical,arslan2025surrogate,weiner2025model,liu2025model,ye2025model,tao2025control}. In low-dimensional or weakly nonlinear settings, such as laminar separation control or simplified wake stabilization, it is competitive and can require orders of magnitude fewer interactions than model-free learning\cite{mao2024model}. This matters most where each evaluation is costly, as in high-fidelity simulation and experiment, and high-speed flows are the clearest case. Shock-resolving compressible solvers are expensive, and hypersonic simulations are prohibitive for the many episodes that model-free training requires. The most direct high-speed example is the control of hypersonic shock-wave/laminar-boundary-layer interaction on a compression ramp with microjets, where a Long Short-Term Memory network trained on CFD data served as the surrogate environment, which reduced training time and allowed the agent to learn quickly\cite{tao2025control}. The CFD data for that surrogate still came from a second-order finite-volume solver, so the surrogate inherits the fidelity of the solver that generated it. By contrast, the model-free alternatives reported so far\cite{fan2020reinforcement,xia2024active} typically pay for their generality with large numbers of expensive solver calls. High-speed model-free agents have been trained directly on the flow solver, including PPO agents for transonic flutter\cite{gong2024active}, a model-free agent for buffet and buffeting in coupled FSI systems\cite{ren2024adaptive}, unstart control for hypersonic flows \cite{mondal2026hypersonic}, and a DRL agent on a fifth-order discontinuous Galerkin environment\cite{mondal2025shocks}. These show that model-free training is feasible at transonic conditions, but the cost scales with solver fidelity.

Model-based ideas also appear in high-speed control outside DRL. Time-variant reduced-order models have been used for adaptive buffet control\cite{gao2025transonic}, and classical closed-loop laws with explicit gains and delays have been used with shock-control bumps\cite{deng2023closed}. These controllers are built on a model of the flow, so they are natural baselines and natural scaffolds for model-based DRL.

\vspace{0.2cm}
\noindent \textbf{Where it breaks down}:
The advantage fades as flow complexity grows. Fully turbulent flows involve strong nonlinearity and interactions across many scales; learning an accurate predictive model of such a system is hard, and small model errors compound over multi-step rollouts, degrading planning in high-Reynolds-number and three-dimensional flows\cite{janner2019trust,chua2018deep}. High-speed flows add difficulties of their own. Shocks are discontinuities whose position and strength change with small shifts in Mach number or angle of attack, shock--boundary layer interaction couples unsteady separation to shock motion, and in FSI problems the structure feeds back on the flow. Smooth learned models may represent such features poorly, and a surrogate fitted to a limited range of conditions may be unreliable outside it. The high dimensionality of the state compounds the problem. Practical implementations therefore compress the flow with proper orthogonal decomposition or autoencoder-based latent representations\cite{brunton2020data,hafner2020mastering}, at the risk of discarding small-scale features that matter for control and so limiting robustness to unseen conditions. In compressible flow the discarded content may be the shock itself, since shocks produce broadband content that low-order bases capture poorly.

Model-based methods are also sensitive to uncertainty from turbulence modelling, discretization, sensor noise and incomplete state information. Realistic control typically relies on sparse measurements such as wall pressure or shear stress, which makes the problem partially observable and violates the Markov assumption underlying most RL algorithms\cite{sutton2018reinforcement}. Sparse sensing is the norm at high speed. Deep learning can reconstruct near-wall pressure fluctuations in shock--boundary layer interaction from sparse, under-sampled data\cite{poulinakis2023deep}, and inlet unstart can be predicted from only two to four surface pressure sensors in under 2 ms\cite{min2024machine}, which suggests that learned observers can supply the state information that a model-based agent lacks. Recurrent networks such as the LSTM surrogate above\cite{tao2025control} also provide memory that partly compensates for partial observability. When the learned model departs from the true dynamics, particularly under changing operating conditions, policies can become suboptimal or unstable, and tend to be less robust than model-free or hybrid alternatives outside their training distribution\cite{moerland2023model}. This matters where failure is costly, such as flutter or inlet unstart. Ongoing efforts address this through uncertainty-aware ensembles\cite{chua2018deep}, physics-informed learning that embeds conservation laws\cite{raissi2019physics} and hybrid model-based and model-free schemes. A high-speed example of a hybrid is the combination of active disturbance rejection control with a TD3 agent for scramjet fuel pulse injection\cite{liang2025research}, where the classical loop handles disturbances and the learned part adjusts injection parameters to meet thrust demands and avoid unstart. Model-based DRL is unlikely to displace model-free learning in highly turbulent regimes, but it remains a valuable component of data-efficient, interpretable control. For high-speed flows, a practical division may be a model-free agent trained on a high-order solver for validation, with a surrogate or reduced-order model used for cheap pretraining and a classical controller as supervisor.

\vspace{0.2cm}
\noindent \textbf{Foundation models as a way forward?}:
Scientific foundation models (SciFMs)\cite{menon2026scientific} offer a different route. Rather than learning task-specific dynamics for a single regime, they aim to capture transferable representations of fluid behaviour from large, heterogeneous datasets spanning configurations, Reynolds/Mach numbers and actuation strategies. Used less as explicit predictors for planning than as sources of latent representations or physical priors, they could reduce sensitivity to local modeling errors, improve robustness under distribution shift and allow rapid adaptation to new objectives with little extra data. For high-speed flows, a model pretrained across Mach numbers, airfoils and ramp or isolator geometries could in principle cut the cost of each new configuration, which is attractive where a single high-order or hypersonic simulation is already expensive. \textit{\textbf{Their central limitation, however, is numerical fidelity: current deep-learning-based SciFMs do not deliver the high order of accuracy of traditional solvers.}} Unlike high-order schemes, whose error decreases at a known rate under grid and time-step refinement, learned models have no formal order of accuracy and no guarantee of convergence. Their errors are typically set by training data and architecture rather than by discretization, and they tend to damp small scales and high-frequency content. These are precisely the near-wall structures, pressure fluctuations and low-frequency shock motion on which control objectives depend. For this reason, SciFMs cannot yet replace high-order solvers as the environment in which policies are trained and certified, and are better suited to supplying priors or accelerating them, with the solver acting as the reference for accuracy. Shocks sharpen the concern, since representations learned from largely smooth data may smear discontinuities or misplace them, and the resulting errors can be exploited by an agent in the same way as numerical artefacts. SciFMs also bring other difficulties: the cost of offline pretraining and online fine-tuning, the reduced interpretability of learned representations and open questions about stability when embedded in a closed loop. They may therefore complement model-based control and high-order simulation by shifting emphasis from precise local prediction to transferable physical understanding, though this remains a nascent direction.

\section{Choosing an algorithm}
Algorithm choice shapes both learning efficiency and control robustness in flows that are nonlinear, high-dimensional and often chaotic (Table~2). Model-free methods used in flow control fall into two broad families, \textit{on-policy} and \textit{off-policy}, and the distinction matters because it sets how much data each interaction with the flow is worth.

On-policy policy-gradient methods, notably proximal policy optimization (PPO)\cite{schulman2017ppo}, are the most widely used for their stability and ease of implementation. The clipped objective limits the size of each policy update, which reduces the chance of catastrophic degradation, an advantage in closed-loop control where an unstable policy can excite undesirable flow modes. PPO underlies the early demonstrations of vortex-shedding suppression behind a cylinder\cite{rabault2019artificial}, actuator-selection strategies\cite{paris2023reinforcement} and multi-agent control of three-dimensional cylinder flows\cite{suarez2025flow}. The price is low sample efficiency, because each batch of data is discarded after a few updates, which becomes a bottleneck when simulations or experiments are expensive\cite{sutton2018reinforcement}.

\begin{table}[t]
\centering
\small
\renewcommand{\arraystretch}{1.25}
\caption{\textbf{Representative model-free DRL algorithms for flow control}}
\label{tab:algos}
\begin{tabularx}{\textwidth}{@{}>{\raggedright\bfseries}p{16mm} >{\raggedright\arraybackslash}p{27mm} >{\raggedright\arraybackslash}X >{\raggedright\arraybackslash}X >{\raggedright\arraybackslash}p{30mm}@{}}
\toprule
Algorithm & Class & Strengths & Limitations & Flow-control examples \\
\midrule
PPO \cite{schulman2017ppo} & On-policy, policy gradient & Stable training through clipped updates; easy to implement; dependable baseline & Low sample efficiency, costly with expensive simulations or experiments & \cite{rabault2019artificial,paris2023reinforcement,suarez2025flow} \\
DQN \cite{mnih2015human} & Off-policy, value-based & Reuses experience through a replay buffer; simple for discrete actuation (e.g.\ on/off jets) & Restricted to discrete actions; overestimates action values & \cite{shimomura2020closed} \\
DDPG \cite{lillicrap2016ddpg} & Off-policy, deterministic actor--critic & Continuous actions; improved sample efficiency & Sensitive to hyperparameters; brittle under turbulent or noisy conditions & \cite{fan2020reinforcement,guastoni2023deep} \\
TD3 \cite{fujimoto2018addressing} & Off-policy, deterministic actor--critic & Twin critics and delayed updates curb overestimation; more stable than DDPG & Still deterministic, so exploration relies on added noise & \cite{liang2025research} \\
SAC \cite{haarnoja2018sac} & Off-policy, entropy-regularized & Sustained exploration; smoother control laws; balance of efficiency and robustness; suited to multistable flows & Comparative evidence on turbulent benchmarks still to be established & \cite{mondal2026hypersonic} \\
\bottomrule
\end{tabularx}
\end{table}

Off-policy algorithms reuse past experience and thereby improve sample efficiency, which is valuable in flow control because collecting new trajectories can require long integration times to obtain statistically converged responses. The deep Q-network (DQN)\cite{mnih2015human} was the first of this family applied to flow control, for example in closed-loop separation control over an airfoil\cite{shimomura2020closed}, but it is limited to discrete actions and therefore to coarse actuation. Deep deterministic policy gradient (DDPG)\cite{lillicrap2016ddpg} extends off-policy learning to continuous actuation and has been used for bluff-body control in experiments and simulations\cite{fan2020reinforcement} and for drag reduction in turbulent channel flow\cite{guastoni2023deep}. However, it couples a deterministic policy with strong sensitivity to hyperparameters, which can produce brittle learning dynamics and reduced robustness in turbulent or noisy conditions\cite{haarnoja2018sac}. Twin delayed DDPG (TD3)\cite{fujimoto2018addressing} mitigates the overestimation of action values that destabilizes DDPG by using two critics and delayed policy updates; it has been adopted, for example, for scramjet fuel injection control\cite{liang2025research}.

Soft actor--critic (SAC)\cite{haarnoja2018sac} maximizes both expected return and policy entropy, which sustains exploration and mitigates premature convergence. This is particularly relevant in multistable flows, where several attractors can coexist and greedy exploitation may trap the controller in a locally good but globally inefficient state. \textit{Entropy regularization} also tends to give smoother control laws that are less sensitive to measurement noise, and off-policy learning makes efficient use of the sparse, delayed rewards typical of drag- or lift-based objectives. These properties suit high-dimensional, partially observable flows, and SAC currently appears to offer the most favourable balance of efficiency and robustness, whereas PPO remains a dependable baseline. Systematic, like-for-like comparisons on turbulent benchmarks are still needed to confirm this ranking as the field moves from low-dimensional test cases to realistic, high-Reynolds-number problems.
\section{Rewards shape what is learned}
In DRL the reward is the only feedback, so its design controls the speed of convergence, the robustness of the policy and the physical relevance of the strategy that emerges\cite{sutton2018reinforcement,silver2021reward}. Naive rewards, such as instantaneous drag or lift, can be sparse or noisy and lead to unstable updates. Shaping the reward with denser signals, penalties on actuation energy or multi-objective formulations can accelerate learning and discourage non-physical shortcuts\cite{rabault2019artificial,van2016deep}. Shaping must be applied with care, because a poorly chosen term can bias the agent towards suboptimal behaviour or hide the essential dynamics\cite{ng1999policy}.

Learned flow-control policies are often highly sensitive to such choices: modest changes in the weights on drag, energy cost or vorticity-related metrics can produce qualitatively different actuation\cite{rabault2019artificial}. Domain knowledge is therefore not optional. Physics-informed rewards that encode symmetries, invariants or conservation laws steer the agent towards strategies that are effective and physically interpretable\cite{raissi2019physics,wan2021rlcontrol,banerjee2025survey}. Penalizing violations of mass or momentum conservation, for example, can prevent exploitation of unrealistic flow manipulations, and emphasizing the suppression of coherent structures can focus control on dynamically relevant features. More broadly, the view that a well-specified reward can in principle give rise to sophisticated behaviour\cite{silver2021reward} suggests that rewards should capture cumulative flow performance, balance competing objectives such as drag reduction against actuation cost, and include regularization that prevents oscillatory actuation. In practice, reward design is an iterative process that combines domain expertise, empirical testing and guidance from RL theory.

\section{Sensing and sampling}
\vspace{0.2cm}
\noindent \textbf{Sparse sensing}:
Practical AFC is constrained by sparse, imperfect sensing. DRL controllers can nevertheless reach effective performance with a small number of wall-mounted sensors, provided these capture the dominant dynamics such as separation or vortex shedding, and the loss relative to full-state feedback is often modest. The minimum sensing requirement is problem-dependent and tied to the dimensionality of the underlying dynamics. Flows dominated by a few coherent structures may need only a handful of well-placed probes, whereas broadband turbulence and multi-scale interactions demand richer sensing. Placement in regions of high flow sensitivity matters more than sensor count, and placement can itself be optimized within the DRL framework\cite{Paris2021}.

\textit{Noise}, \textit{delay} and \textit{partial observability} are the main threats. Noise can bury weak but dynamically important signals, delays make the controller act on outdated information and may destabilize the loop, and policies trained under idealized sensing often lose robustness when exposed to realistic noise and latency. Recurrent architectures and temporal state augmentation can partly restore the Markov property by encoding flow history, and robust training together with observer-based hybrids will be important for deployment.

\vspace{0.2cm}
\noindent \textbf{How often to sample}:
The sampling rate must resolve the dominant time scales of the flow, such as vortex shedding, shear-layer oscillations or energy-containing eddies\cite{pope2000turbulent,Paris2021}. A practical analogue of the Nyquist criterion requires
\begin{equation}
f_s \gtrsim 2 f_{\max},
\end{equation}
where $f_{\max}$ is the highest relevant flow frequency. Undersampling causes aliasing, masks the causal link between state and action, and can yield policies that act out of phase with the flow or excite unintended dynamics\cite{bewley2018feedback}. The flow does not alone set the optimum, however, because actuator bandwidth $f_\mathrm{act}$ limits how fast the system can respond, and additional temporal resolution beyond it is wasted. A physically informed choice aligns both:
\begin{equation}
f_\mathrm{opt}\sim\min\left(2f_{\max},\,f_\mathrm{act}\right).
\end{equation}
Spatial sensing interacts with this choice: sparse sensors may call for higher temporal rates to preserve observability, whereas dense networks can tolerate lower rates when spatial information is rich enough to reconstruct the flow\cite{rudin1991inverse}. Oversampling adds little and carries costs, including more data, longer episode horizons that slow training and redundant correlations that complicate function approximation\cite{sutton2018reinforcement}. Event-triggered and adaptive sampling have been proposed to balance resolution and efficiency\cite{furst2019event}.

\section{Explainable DRL}
\begin{figure}[htpb]
\centering
\includegraphics[trim=2cm 1cm 0.4cm 0cm, scale=0.44, clip=true]{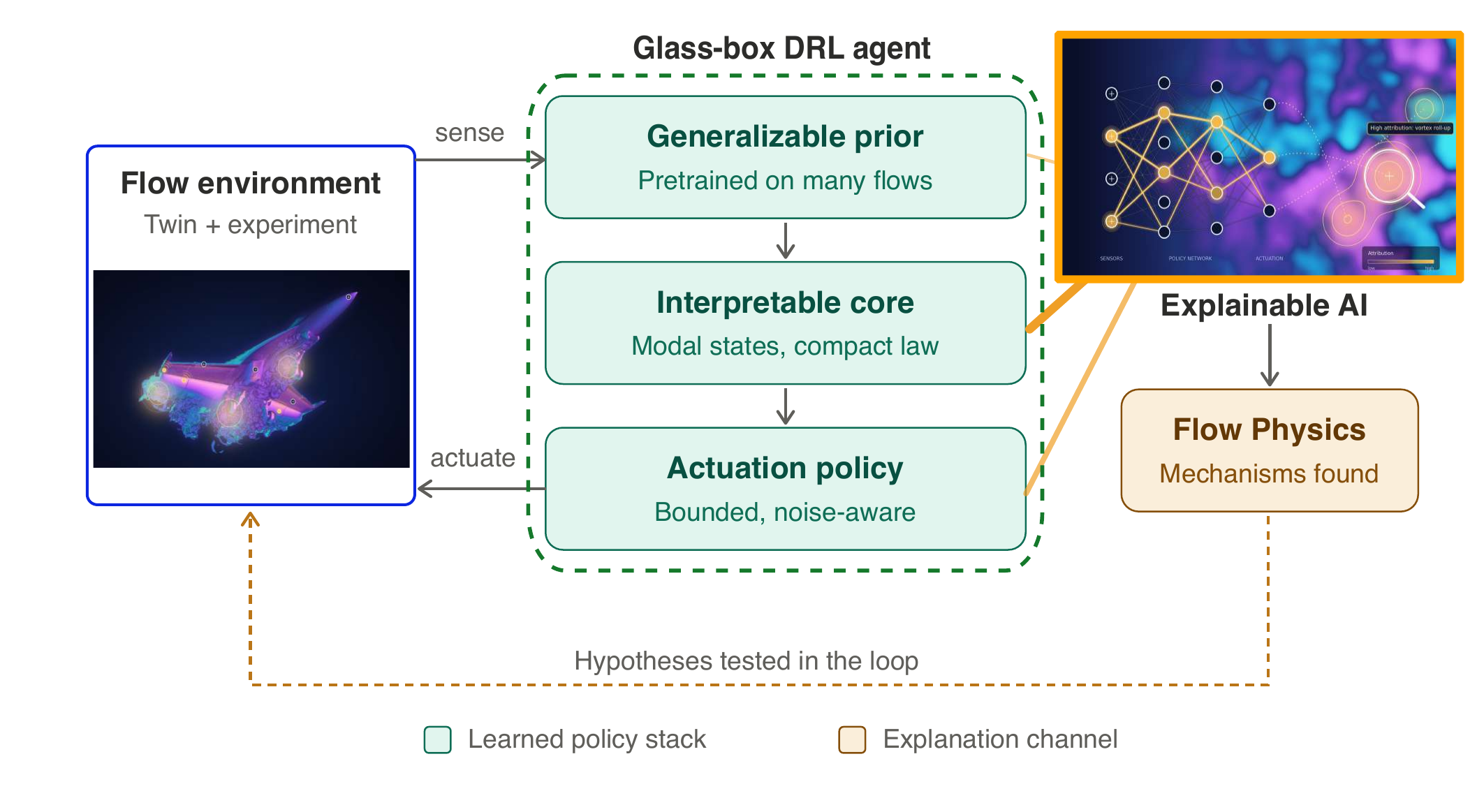}
\caption{\textbf{Glass-box DRL policy}: Post hoc explanation (attribution, embeddings and interventions) links sensed inputs and actuation to the flow mechanisms the policy exploits, such as shock motion, separation and fluid--structure coupling.}
\label{fig:F2}
\end{figure}

A DRL policy is a high-dimensional function fitted to a reward, and high reward alone is a weak basis for trust. The need for more is sharpest in high-speed flows, where failure is costly: flutter can damage a structure, and unstart can cut thrust in a scramjet. A controller that is to be flown, certified or built upon must be explained: which measurements does it use, what does it do with them, and which flow mechanism does that action exploit? Explainability, in the sense used here, is the post hoc analysis of a trained policy that is treated as a black box (Box~2). It asks for an account of behaviour that already exists, without changing the policy itself. Figure~\ref{fig:F2} sketches the resulting glass-box view, in which attribution, embedding and intervention tools open the policy to inspection and tie its behaviour to flow physics.

\begin{naturebox}{Box 2 $|$ Explainable, interpretable and generalizable DRL}
\textbf{Explainable} policies are analysed after training to account for their behaviour, using attribution, embeddings or interventions. The policy itself is unchanged and may remain a black box.\vspace{2pt}

\textbf{Interpretable} policies are transparent by construction. Their form is restricted so that inputs, structure and parameters have direct physical meaning.\vspace{2pt}

\textbf{Generalizable} policies retain performance when conditions differ from those seen in training, such as Reynolds or Mach number, geometry, shock position, sensing and actuation layout, or the move from simulation to experiment.\vspace{2pt}

The three properties are linked. An explanation can reveal which features a policy relies on, an interpretable form can force reliance on features that are physically invariant, and invariant features are the ones most likely to transfer.
\end{naturebox}

\vspace{0.2cm}
\noindent \textbf{Attribution and sensitivity}:
The most direct tools quantify how much each input drives each decision. Saliency maps and feature-importance scores, developed for visual RL agents\cite{greydanus2018visualizing}, translate naturally to AFC, where inputs are sensor signals or flow-field snapshots and the question becomes which probes, which regions of the field and which time lags the policy relies on. In high-speed flows the inputs are typically a few wall-pressure or force signals, because full-field data are rarely available in experiments. Sparse-sensing studies show that a handful of surface pressure sensors can suffice to reconstruct shock--boundary layer interaction loads\cite{poulinakis2023deep} or to predict inlet unstart\cite{min2024machine}, and DRL controllers with sparse surface pressure feedback have been developed for incompressible bluff-body flow\cite{wang2024dynamic}. Attribution then indicates which of those few sensors matter and at what delay. It is most informative when compared with flow physics. If the highest-attribution sensors sit near the shock foot, in the separation bubble or on the surface swept by the oscillating shock, the policy is attending to the instability it is meant to control, as in transonic buffet\cite{ren2024adaptive} or shock--boundary layer interaction\cite{tao2025control}. If they sit where the flow carries no relevant information, the policy may be exploiting an artefact of the training environment.

\vspace{0.2cm}
\noindent \textbf{Embeddings of policy behaviour}:
Dimensionality reduction of the state--action manifold, using principal component analysis or t-distributed stochastic neighbour embedding\cite{hinton2008visualizing}, can reveal whether a policy occupies a small number of distinct actuation modes, switches between them, or varies continuously with flow phase. Coloured by flow quantities such as shock position, buffet phase, pitch and plunge phase in flutter\cite{gong2024active}, separation-bubble size or thrust, these embeddings link regions of policy space to coherent structures such as shocks, vortices and shear layers, which gives an interpretable vocabulary for a policy that is otherwise numeric. Modal decompositions of controlled and uncontrolled flows\cite{brunton2020data} complement this view from the flow side, showing which structures the controller amplifies or suppresses. In early cylinder-wake studies, for example, learned actuation correlated with phase-locked forcing of vortex shedding, which points to implicit exploitation of the underlying instability\cite{rabault2019artificial}, although the policy itself revealed little about how it did so without such targeted analysis. The analogous question for shocked flows is whether an agent locks its actuation to the shock oscillation or to a structural frequency, and whether this is the mechanism that extends a flutter boundary beyond the training conditions\cite{gong2024active}.

\vspace{0.2cm}
\noindent \textbf{Intervention and causal analysis}:
Correlation is not mechanism. Because the DRL loop can be simulated, explanation can be interventional: individual sensor channels can be masked or perturbed, actuation can be clamped or time-shifted, and the response of both policy and flow observed. This separates sensed features that cause an action from those that merely co-vary with it, and actions that change the flow from those that do not\cite{pearl2009causality}. In a closed loop the distinction is subtle, because the policy changes the flow that it then senses, so interventions must be designed with the feedback in mind. High-speed flows add specific tests. Shifting the actuation phase relative to the shock oscillation, silencing the sensor nearest the shock, or changing the Mach number or angle of attack slightly shows whether the policy tracks the shock or only a correlated signal. These interventions also allow a direct comparison with classical controllers such as lift-feedback shock-control bumps\cite{deng2023closed} or reduced-order-model-based control\cite{gao2025transonic}, whose gains and delays are explicit, and with actuation concepts such as boundary-layer suction\cite{sethuraman2021control} or electrical discharge\cite{hahn2025control} that act on the shock structure directly. If a learned policy reproduces the logic of such a controller, that is evidence of physical grounding. If it departs from it, the departure is either a discovery or a warning.

Explanations are models of a policy, not the policy, and can be unfaithful, so faithfulness should be tested: removing sensors marked important should degrade control, and removing those marked irrelevant should not. Explanation also audits for numerical artefacts that agents exploit. In compressible flows, an agent trained on a low-order solver may be tuned to shock smearing rather than shock dynamics, whereas high-order solver resolve shock oscillations more sharply\cite{mondal2026hypersonic,mondal2025shocks}. Learned surrogates, which can cut hypersonic training cost\cite{tao2025control}, pose the same risk, since explanations may describe the surrogate’s errors rather than the flow. Re-running attribution and intervention tests on the full-order solver, and ultimately on experiments, exposes such failures before a policy is trusted for tasks such as unstart prevention\cite{liang2025research}.

\section{Interpretable DRL}
Where explainability analyses a finished policy, interpretability builds transparency in. An interpretable policy is one whose structure can be read directly: its inputs, functional form and parameters carry physical meaning, so that its behaviour follows from its design rather than from an after-the-fact account (Box~2). This is the stronger property and the harder one to obtain, because it constrains the very flexibility that makes DRL attractive. The constraint is worth accepting in high-speed flows, where the controller must hold up in regimes that cannot be flight-tested cheaply, and where a policy whose logic is open to inspection is far easier to certify than one that is not.

\vspace{0.2cm}
\noindent \textbf{Controlling the policy class}:
The simplest route is to restrict what the policy can express. Small networks with few inputs, bounded actuation and smoothness regularization yield control laws that can be inspected and bounded, and they reduce sensitivity to measurement noise. This suits high-speed problems, where sensing is naturally sparse, such as a few surface pressure probes\cite{min2024machine,poulinakis2023deep}, and actuation is naturally bounded, such as synthetic jets\cite{gong2024active} or microjets\cite{tao2025control}. Structured policies go further by building known invariants or symmetries into the architecture, so that the admissible laws are consistent with the physics from the start\cite{raissi2019physics}. Examples include non-dimensional inputs that make a policy portable across Mach number, or actuation constrained to respect the phase relation between pitch and plunge. This is attractive wherever an unconstrained network would have to rediscover such structure from data.

\vspace{0.2cm}
\noindent \textbf{Kolmogorov--Arnold networks}:
Interpretability can also be sought in the network architecture itself. Kolmogorov--Arnold networks (KANs)\cite{liu2025kan} replace the fixed activations and linear weights of a multilayer perceptron with learnable univariate functions on the edges, so that each connection can be plotted, pruned and, where the learned shape is simple, replaced by a closed-form expression. A KAN policy, or a KAN critic, can therefore be read as a composition of one-dimensional functions that relate each sensed quantity to the actuation, which makes it a natural bridge between neural and symbolic control laws. For high-speed flows this means a policy that maps wall-pressure signals to jet amplitude through curves that can be compared directly with the gain and delay of a classical shock-control law\cite{deng2023closed}. KANs have been compared with multilayer perceptrons on scientific problems such as differential equations and operator learning\cite{menon2026fekan}, but their use as DRL policies for compressible flow has barely been explored. Open questions include their sample efficiency and cost inside a costly CFD loop, their behaviour under noisy sensing, and whether the learned edge functions stay simple enough to interpret when the dynamics involve discontinuities such as shocks.

\vspace{0.2cm}
\noindent \textbf{Learning in physical coordinates}:
A further route changes the space in which the policy acts. Coupling DRL to reduced-order models built from proper orthogonal decomposition (POD), dynamic mode decomposition (DMD) or resolvent analysis lets the agent act on a handful of modal amplitudes instead of raw sensor data, and projects its strategy onto physically meaningful bases\cite{brunton2020data,liu2025model}. \textit{This route is poorly suited to high-speed flows in which shocks appear.} POD and DMD represent the flow as a superposition of global, smooth modes, whereas a shock is a near-discontinuity whose spectral content extends to ever higher wavenumbers, tending to infinity in the inviscid limit. A low-order basis therefore converges slowly, needs many modes to resolve even a single shock, and tends to smear the shock or produce spurious oscillations around it. The difficulty grows when the shock moves, because a translating discontinuity is not well described by a linear combination of fixed modes, so the truncated model loses the very feature that controls loads, separation and unstart. In high-speed flows the natural coordinates are well known: the shock position and the buffet mode in transonic flow, pitch and plunge in flutter, the leading-edge position of the shock train in an isolator\cite{sethuraman2021control,hahn2025control}, and the back-pressure ratio that governs inlet unstart\cite{mondal2026hypersonic,min2024machine}. The state is then a set of named flow structures, and the policy is a map between them that can be examined directly. Time-variant reduced-order models have already been used for adaptive buffet control\cite{gao2025transonic}, and pairing such models with a DRL agent is a natural next step, provided the shock is tracked explicitly through such coordinates rather than left to a modal expansion. The price is that anything outside the retained modes is invisible to the agent, which matters when shocks produce broadband content that low-order bases capture poorly.

\vspace{0.2cm}
\noindent \textbf{Distillation and hybrid structure}:
A fourth route compresses a trained policy into a compact, human-readable form, such as a low-order symbolic expression or a shallow decision structure, and verifies that the compact law retains most of the performance. Programmatically interpretable policies\cite{verma2018programmatically}, decision-tree extraction\cite{bastani2018verifiable} and sparse regression of governing or control laws\cite{brunton2016discovering} provide the tools, and a distilled law for buffet or flutter suppression would be directly comparable with the closed-loop laws used in classical practice\cite{deng2023closed}. Hybrid designs reach a similar goal from the other direction, using a classical controller such as a linear quadratic regulator or model predictive control as the scaffold and letting DRL learn only a bounded correction\cite{brunton2015closed}. The combination of active disturbance rejection control with a TD3 agent for scramjet fuel injection\cite{liang2025research} is an example from the high-speed literature: the classical part is interpretable and certifiable, and the learned part is confined to where it adds value. We return to such hybrids below.

\vspace{0.2cm}
\noindent \textbf{Rewards and constraints as interpretable specifications}:
The objective itself can be made interpretable. Physics-informed rewards that penalize actuation energy, violation of conservation laws or non-physical manipulation\cite{wan2021rlcontrol,banerjee2025survey} state what the controller is meant to do in terms that engineers can audit. In high-speed settings the specification is often explicit already: drag reduction with or without a lift penalty\cite{mondal2025shocks}, thrust demands together with unstart avoidance\cite{mondal2026hypersonic,liang2025research}, or bounds on pitching and plunging amplitude\cite{gong2024active}. Constrained or safe RL formulations make actuator limits, energy budgets and safety margins explicit rather than implicit in reward weights, so that a hard boundary such as the unstart limit is enforced by construction instead of being traded against thrust.

Each of these routes gives something up. A restricted policy class cannot represent every nonlinear strategy that an unconstrained network can, and the very behaviours that make DRL interesting, such as exploitation of subtle instabilities or cross-scale interactions, may be the ones a constrained form cannot express. Compressible flows sharpen the trade-off, because shock dynamics are strongly nonlinear and a policy that is too simple may miss the nonlinear shock--boundary layer coupling it is meant to exploit. We therefore expect a \textit{Pareto frontier} rather than a single answer, and the field would benefit from reporting both control performance and a measure of policy complexity. The payoff is more than trust. An interpretable policy turns learned behaviour into a hypothesis that conventional analysis can test, which is how DRL can contribute to flow physics rather than only to flow engineering.

\section{Generalizable DRL}
Generalization is where the promise of DRL is most exposed, and high-speed flows expose it further. Policies are trained at a fixed Reynolds number and Mach number, on a fixed geometry, with a fixed sensor and actuator layout, in an idealized solver, and are then expected to work elsewhere. In practice, \textit{zero-shot transfer} is limited. A policy exploits the coherent structures, dominant frequencies and instability mechanisms of its training regime, and even moderate changes in operating conditions alter the spectrum and the balance of the governing mechanisms enough to degrade control rapidly. Controllers trained to suppress vortex shedding are tuned to a narrow band of Strouhal numbers and wake dynamics\cite{rabault2019artificial,fan2020reinforcement}. In compressible flow the dependence is at least as strong, because small changes in Mach number or angle of attack move the shock, shift the buffet onset boundary or change the flutter margin, and the mapping from state to action changes with them. Standard agents thus appear to learn \emph{environment-specific} strategies rather than physics-invariant laws. The few high-speed studies that report generalization are therefore notable. A DRL flutter controller has been reported to reduce pitching and plunging oscillations across a range of Mach numbers and to extend the flutter boundary beyond the training conditions\cite{gong2024active}, and a machine-learning unstart predictor has been designed for a wide speed range with only a few surface pressure sensors\cite{mondal2026hypersonic,min2024machine}. Such results show that transfer is possible, but each covers one axis over a stated range, and none shows that the policy has learned an invariant law.

\vspace{0.2cm}
\noindent \textbf{Axes of generalization}:
It helps to separate what is being generalized over. Parametric generalization concerns Reynolds number, Mach number, angle of attack and operating conditions. For a transonic airfoil this includes the lift-penalty weight in the reward, since drag reduction with and without a lift penalty are different tasks\cite{mondal2025shocks}. Geometric generalization concerns body shape, boundary conditions and the location of actuators and sensors, for example a different airfoil section, a different isolator or ramp geometry, or the placement of jets and pressure probes. Regime generalization concerns the move between laminar, transitional and turbulent states, and between compressible regimes where the dominant mechanisms change, from attached transonic flow to buffet, from supersonic to hypersonic shock--boundary layer interaction, or from started to unstarted inlet operation. Finally, generalization from simulation to experiment, taken up in the next section, concerns the fidelity gap, sensor noise, latency and actuator limits. A policy can generalize well along one axis and fail along another, so claims of generalization should name the axis and the range over which it was tested.

\vspace{0.2cm}
\noindent \textbf{Training for diversity}:
\emph{Domain randomization} varies key parameters during training so that the policy must rely on features common to many flow realizations\cite{Tobin2017}. For high-speed flows the natural parameters are Mach number, angle of attack, structural stiffness and damping in FSI problems, and the shock position at the start of an episode. It does not guarantee transfer, but it enlarges the range of conditions with acceptable performance, and randomizing noise, latency and actuator response begins to address the simulation-to-experiment axis. \emph{Multi-fidelity and curriculum training} begin with cheap, low-fidelity or low-Reynolds-number data and refine on expensive data, aided by faster solvers\cite{kochkov2021machine}. This is attractive when a hypersonic or high-order simulation is prohibitive, and learned surrogates already serve as cheap training environments for hypersonic control\cite{tao2025control}. Their benefit is plausible but not yet established for flow control, and the central risk is importing the biases of the cheap model. In compressible flow this risk is concrete: a surrogate fitted to a limited range of conditions may be unreliable outside it, and a low-order solver may smear the shock and alter its oscillation, whereas high-order schemes resolve it more sharply\cite{mondal2025shocks}. A policy trained on the cheap model can thus be tuned to its errors, which is a transfer failure in its own right.

\vspace{0.2cm}
\noindent \textbf{Training to adapt}:
\emph{Transfer learning} pretrains a policy in a simpler setting and fine-tunes it in a harder one, reusing control-relevant representations such as separation detection or shear-layer manipulation, and so cuts training cost\cite{yan2025deep}. Transfer from two- to three-dimensional bluff-body flow has been demonstrated\cite{he2023policy,yan2025deep}, and the same logic suggests pretraining on two-dimensional airfoil buffet or shock--boundary layer interaction before fine-tuning on three-dimensional wings or ramps, where simulations are far more expensive. \emph{Meta-learning}\cite{Finn2017} and multi-task RL\cite{wilson2007multi,sodhani2021multi} optimize explicitly for fast adaptation, which introduces an inductive bias towards generalization across flow and geometry parameters. Applications to fluid mechanics are at an early stage, but they point to controllers that adapt with a few interactions instead of being retrained from scratch, for example, across a family of Mach numbers or after a change in structural properties.

\vspace{0.2cm}
\noindent \textbf{Learning what is invariant}:
The deepest route is to make policies depend on what does not change. Representations that encode physical structure, whether built in through symmetries and invariants\cite{raissi2019physics}, obtained from reduced-order bases\cite{brunton2020data}, or pretrained as latent priors in scientific foundation models\cite{menon2026scientific}, should transfer further than raw-sensor policies because they discard condition-specific detail. In high-speed flows, non-dimensional inputs and physically named coordinates are the obvious candidates: shock position relative to the chord, pitch and plunge amplitudes, or the back-pressure ratio that governs unstart\cite{mondal2026hypersonic,min2024machine}. Reduced-order models that vary in time have already supported adaptive buffet control across changing conditions\cite{gao2025transonic}, and a DRL agent acting on such coordinates is a natural next step. Here generalization meets interpretability: a policy that is shown, by explanation or by construction, to rely on invariant features is a better candidate for transfer than one that is not. Classical laws offer a useful benchmark, since lift-feedback shock control\cite{deng2023closed} works through explicit gains and delays that carry across conditions by design. Foundation models carry their own costs, including pretraining and fine-tuning expense and uncertain stability in closed loop, and shocks, which are discontinuous, may be hard for smooth latent representations to capture.

Progress needs shared protocols. Policies should be evaluated on held-out Reynolds numbers, Mach numbers, angles of attack, geometries and sensor layouts rather than only on the training condition, with zero-shot and few-shot performance reported separately and with robustness to noise and delay tested explicitly. For high-speed flows the test should also include the regime boundaries, such as conditions near buffet onset, the flutter boundary or the unstart limit, because these are where a failure to generalize is most costly and where a policy trained deep inside a regime is most likely to fail. Without such benchmarks, improvements in learning efficiency are easy to mistake for improvements in generalization. The measure that ultimately matters for deployment is not how well a policy performs where it was trained, but how gracefully it degrades where it was not.
\section{From simulation to experiment}
Moving from simulation to experiments or real systems is the hardest step, and in high-speed flows it is harder still. The central obstacle is the simulation-to-reality gap (Fig.~\ref{fig:loop}): high-fidelity simulations rely on idealized boundary conditions, perfect state observability and deterministic dynamics, assumptions that rarely hold in a laboratory\cite{brunton2015closed,gautier2015closed}. Policies trained in simulation may exploit numerical artefacts or control authority that cannot be realized, and become unstable or suboptimal when deployed. The DRL studies of transonic buffet, flutter and drag reduction\cite{ren2024adaptive,gong2024active,mondal2025shocks}, hypersonic shock-wave/boundary-layer interaction\cite{tao2025control,mondal2026hypersonic} and scramjet fuel injection\cite{liang2025research} are numerical, so none has yet crossed this gap. Experimental work on shock control does exist, such as the stabilization of Mach 2 shock trains with electrical discharges\cite{hahn2025control}, but it uses fixed actuation rather than a learned policy. Zong et al.\cite{zong2025closed} ran DRL on a field-programmable gate array (FPGA), achieving 1--10 kHz control, two orders of magnitude faster than CPU-based frameworks. In Mach 2 supersonic backward-facing-step flow, the agent learned an effective control law within about ten seconds, increasing shear-layer mixing by 21.2\%. 

Closing the distance between the two is a central task for the field. High-speed facilities add their own constraints: run times are often short, test conditions are costly to reproduce, and a failure such as unstart or structural damage can end an experiment or destroy a model. This limits the trial-and-error learning on which DRL depends, so much of the training must still happen in simulation, and the gap matters more.

\vspace{0.2cm}
\noindent \textbf{Modelling error, noise, latency and actuation uncertainty}:
Modelling errors change the perceived dynamics and can invalidate the state--action relationships learned in training\cite{ye2025model}. In compressible flow the error often sits in the shock. Shock position, oscillation amplitude and frequency depend on the shock-capturing scheme, limiters, numerical dissipation and turbulence modelling, so a policy trained on a low-order solver may be tuned to the smearing of the discontinuity rather than to its physical dynamics, whereas high-order schemes resolve shock oscillations more sharply\cite{mondal2025shocks}. Surrogate environments add another layer, because a policy trained on a learned model of the flow inherits that model's errors, and the CFD data behind it carry the fidelity of the solver that produced them\cite{tao2025control}. In FSI problems the structural model is a further source of mismatch, since stiffness, damping and unmodelled nonlinearities in a real structure differ from those assumed in simulation\cite{ren2024adaptive,gong2024active}.

Sensor noise corrupts observations and effectively turns a fully observable decision process into a partially observable one, for which standard algorithms are not designed. High-speed experiments rarely offer more than a few wall-pressure signals, which are affected by noise, vibration and thermal effects. Sparse-sensing studies show that useful information can be extracted from such data: near-wall pressure fluctuations in shock--boundary layer interaction can be reconstructed from sparse, under-sampled measurements\cite{poulinakis2023deep}, and inlet unstart can be predicted from only two to four surface pressure sensors\cite{min2024machine}. These results suggest that learned observers could supply the state estimate for a DRL controller, although neither study tests this. Noise-aware training, in which the same sparse, noisy signals available in the experiment are used as the agent's observations, is a direct way to prepare for the sensing conditions of deployment.

Latency in sensing, computation and actuation introduces delays that can destabilize closed-loop control, especially when relevant time scales are short\cite{novati2021automating}. Time scales are short at high speed. Shock oscillation, buffet and inlet unstart evolve quickly, so the whole loop of sensing, inference and actuation must fit inside them, and this is our own estimate rather than a finding of the cited studies. Sub-2-ms prediction of unstart from surface pressure has been reported\cite{mondal2026hypersonic,min2024machine}, which shows that the sensing and inference side can be fast, but actuator response and hardware delays must be added. Actuation uncertainty, including limited bandwidth, saturation, hysteresis and variable response, makes it hard to reproduce the fine-grained strategies learned in simulation\cite{gautier2015closed}. Synthetic jets and microjets, which several of the studies above use\cite{gong2024active,tao2025control}, have finite bandwidth and amplitude limits, and the conditions under which a synthetic jet forms depend on its operating parameters\cite{holman2005formation}, so a policy that assumes ideal jet response may ask for forcing that a real actuator cannot deliver. Electrical-discharge actuation acts on the shock structure directly\cite{hahn2025control}, but its response in a given facility must be characterized before it can be modelled in the training environment. Boundary-layer suction is another candidate, studied numerically for shock trains\cite{sethuraman2021control}.

All together these factors limit real-time performance and robustness, and motivate the domain randomization and adaptation strategies described above, together with noise-aware training, reduced-order sensing and hybrid model-based and learning-based approaches. For high-speed flows, a practical route is to randomize Mach number, structural parameters, sensor noise, latency and actuator response during training, test the policy on a higher-fidelity solver and in hardware-in-the-loop experiments, and keep a classical controller as supervisor to protect against unstart or structural overload. This sequence is our proposal, and it builds on hybrid designs such as active disturbance rejection control combined with a TD3 agent\cite{liang2025research}.

\section{Beyond the efficiency--accuracy trade-off}
AFC is often assumed to force a choice between efficiency and accuracy. High-fidelity simulation and dense sensor--actuator networks are accurate but expensive and may rule out real-time use, whereas low-dimensional models and coarse sampling are cheap but compromise the fidelity of learned policies\cite{brunton2020data,rabault2019artificial}. DRL weakens this dichotomy. Off-policy, entropy-regularized algorithms use experience replay and structured exploration to achieve strong performance with sparse data or reduced-order representations\cite{haarnoja2018sac,wan2021rlcontrol}, and hybrid schemes that couple DRL with reduced-order models or physics-informed surrogates train efficiently while retaining the dynamics that matter\cite{raissi2019physics,kutz2022drlfluids}. Accuracy should in any case be judged not only by instantaneous metrics such as drag reduction, but also by robustness, generalization across Reynolds numbers and interpretability. On those terms the trade-off is a product of conventional modelling constraints, not a fundamental limit.

This is the logic of the hybrid frameworks introduced above. Coupling DRL to reduced-order models gives the agent a low-dimensional latent space that retains the dominant structures, which improves sample efficiency, stabilizes learning and readily accommodates physics-informed rewards, constraints and symmetries\cite{brunton2020data,liu2025model,raissi2019physics,wan2021rlcontrol}. Using classical control as a scaffold confines exploration to a sensible neighbourhood while leaving DRL free to discover nonlinear actuation patterns\cite{rabault2019artificial,brunton2015closed}. What makes hybrids compelling is that the same construction addresses several of the challenges reviewed here at once: it lowers the cost of interaction, supplies interpretable structure, and carries the invariants that help policies transfer across conditions. Hybrid design thus offers one answer to the efficiency, interpretability and generalization problems instead of three separate ones, although how far it can be pushed in fully turbulent, high-speed flows is still open.


\section{Outlook}
DRL has shown that learned controllers can match classical flow control in simulation. Turning these demonstrations into working technology is now a problem of validation more than of algorithms, and it is most acute in high-speed flows. Transonic, supersonic and hypersonic systems combine shock-dominated dynamics that depend on the numerical scheme, costly high-fidelity simulation, sparse sensing, millisecond time scales and failures, such as flutter, buffet and inlet unstart, that are expensive or irreversible. In this setting a black-box policy is a liability. A controller that performs well but cannot say why will not be trusted near the flutter boundary or the unstart limit, and it cannot be certified for flight.

The way forward is glass-box AI: controllers whose behaviour can be inspected, explained and bounded. Interpretability should be built into the policy through physics-informed rewards, hybrid architectures that pair learned components with reduced-order models or classical control, and sparse, symbolic or otherwise compact representations whose terms can be tied to shock motion, boundary-layer separation or pressure recovery. Explainability should be applied after training to identify which sensors, flow features and states drive each action, to check that the policy exploits real flow physics rather than solver-specific artefacts, and to expose its failure modes. Generalizability follows from the same transparency. A policy that encodes the physical mechanism it exploits can be expected to transfer across Mach numbers, angles of attack, geometries and fidelity levels, and when it fails, the failure can be traced to a cause and not merely observed. Each property supports the others: interpretable policies are easier to explain, explainable policies are easier to test for generalization, and policies that generalize for identifiable reasons are easier to certify.

Closing the gap between simulation and flight requires this evidence at every level of technology readiness. Policies must be verified across numerical schemes and fidelity levels, designed around hard structural, thrust and unstart limits, and tested hardest near regime boundaries, then confirmed through hardware-in-the-loop experiments and blowdown or high-enthalpy facilities. Inference must fit within the time scales of the flow using the few sensors a real vehicle can carry, and a classical supervisory layer able to override the agent near the boundary offers a practical route to safety. Progress should be measured against strong classical baselines on open canonical cases, such as a transonic airfoil, a compression ramp and an isolator, with explainability and generalization reported alongside performance. Whether learned controllers are adopted in high-speed flight will depend less on how well they perform than on whether their behaviour can be understood, trusted and verified.

\small
\bibliographystyle{unsrtnat}
\bibliography{ref}

@article{ye2025model,
  title={Model-based reinforcement learning for active flow control},
  author={Ye, Minghui and Elsheikh, Ahmed H},
  journal={Physics of Fluids},
  volume={37},
  number={9},
  year={2025},
  publisher={AIP Publishing}
}

@book{anderson1989hypersonic,
  title={Hypersonic and high temperature gas dynamics},
  author={Anderson, John David},
  year={1989},
  publisher={Aiaa}
}

@techreport{peherstorfer2016survey,
  title={Survey of multifidelity methods in uncertainty propagation, inference},
  author={Peherstorfer, Benjamin and Willcox, Karen and Gunzburger, Max},
  year={2016},
  institution={and optimization. Technical Report TR-16-1. Aerospace Computational Design~…}
}

@article{park1989nonequilibrium,
  title={Nonequilibrium hypersonic aerothermodynamics},
  author={Park, Chul},
  journal={Nonequilibrium hypersonic aerothermodynamics by Park},
  pages={29860},
  year={1989}
}

@article{candler1991computation,
  title={Computation of weakly ionized hypersonic flows in thermochemical nonequilibrium},
  author={Candler, Graham V and MacCormack, Robert W},
  journal={Journal of Thermophysics and heat transfer},
  volume={5},
  number={3},
  pages={266--273},
  year={1991}
}

@article{bertin2003fifty,
  title={Fifty years of hypersonics: where we've been, where we're going},
  author={Bertin, John J and Cummings, Russell M},
  journal={Progress in Aerospace Sciences},
  volume={39},
  number={6-7},
  pages={511--536},
  year={2003},
  publisher={Elsevier}
}

@article{kitamura2013towards,
  title={Towards shock-stable and accurate hypersonic heating computations: A new pressure flux for AUSM-family schemes},
  author={Kitamura, Keiichi and Shima, Eiji},
  journal={Journal of Computational Physics},
  volume={245},
  pages={62--83},
  year={2013},
  publisher={Elsevier}
}

@article{souverein2010effect,
  title={Effect of air jet vortex generators on a shock wave boundary layer interaction},
  author={Souverein, LJ and Debi{\`e}ve, J-F},
  journal={Experiments in Fluids},
  volume={49},
  number={5},
  pages={1053--1064},
  year={2010},
  publisher={Springer}
}

@article{amodei2016concrete,
  title={Concrete problems in AI safety},
  author={Amodei, Dario and Olah, Chris and Steinhardt, Jacob and Christiano, Paul and Schulman, John and Man{\'e}, Dan},
  journal={arXiv preprint arXiv:1606.06565},
  year={2016}
}

@article{mondal2026hypersonic,
  title={Hypersonic flow control: Generalized deep reinforcement learning for hypersonic intake unstart control under uncertainty},
  author={Mondal, Trishit and Jagtap, Ameya D},
  journal={Physics of Fluids},
  volume={38},
  number={4},
  year={2026},
  publisher={AIP Publishing}
}

@article{pandolfi2001numerical,
  title={Numerical instabilities in upwind methods: analysis and cures for the “carbuncle” phenomenon},
  author={Pandolfi, Maurizio and D'Ambrosio, Domenic},
  journal={Journal of Computational Physics},
  volume={166},
  number={2},
  pages={271--301},
  year={2001},
  publisher={Elsevier}
}

@article{quirk1994contribution,
  title={A contribution to the great Riemann solver debate},
  author={Quirk, James J},
  journal={International Journal for numerical methods in fluids},
  volume={18},
  number={6},
  pages={555--574},
  year={1994},
  publisher={Wiley Online Library}
}

@article{clemens2014low,
  title={Low-frequency unsteadiness of shock wave/turbulent boundary layer interactions},
  author={Clemens, Noel T and Narayanaswamy, Venkateswaran},
  journal={Annual Review of Fluid Mechanics},
  volume={46},
  number={1},
  pages={469--492},
  year={2014},
  publisher={Annual Reviews}
}

@article{crouch2009origin,
  title={Origin of transonic buffet on aerofoils},
  author={Crouch, Jeffrey D and Garbaruk, A and Magidov, D and Travin, A},
  journal={Journal of fluid mechanics},
  volume={628},
  pages={357--369},
  year={2009},
  publisher={Cambridge University Press}
}

@article{lee2001self,
  title={Self-sustained shock oscillations on airfoils at transonic speeds},
  author={Lee, BHK},
  journal={Progress in Aerospace Sciences},
  volume={37},
  number={2},
  pages={147--196},
  year={2001},
  publisher={Elsevier}
}

@article{vignon2023recent,
  title={Recent advances in applying deep reinforcement learning for flow control: Perspectives and future directions},
  author={Vignon, Colin and Rabault, Jean and Vinuesa, Ricardo},
  journal={Physics of fluids},
  volume={35},
  number={3},
  year={2023},
  publisher={AIP Publishing}
}

@article{rabault2019artificial,
  title={Artificial neural networks trained through deep reinforcement learning discover control strategies for active flow control},
  author={Rabault, Jean and Kuchta, Miroslav and Jensen, Atle and R{\'e}glade, Ulysse and Cerardi, Nicolas},
  journal={Journal of fluid mechanics},
  volume={865},
  pages={281--302},
  year={2019},
  publisher={Cambridge University Press}
}

@article{wiliamson1996vortex,
  title={Vortex dynamics in the cylinder wake},
  author={Wiliamson, CH},
  year={1996}
}

@article{turkel1999preconditioning,
  title={Preconditioning techniques in computational fluid dynamics},
  author={Turkel, Eli},
  journal={Annual Review of Fluid Mechanics},
  volume={31},
  number={1},
  pages={385--416},
  year={1999},
  publisher={Annual Reviews 4139 El Camino Way, PO Box 10139, Palo Alto, CA 94303-0139, USA}
}

@book{garnier2009large,
  title={Large eddy simulation for compressible flows},
  author={Garnier, Eric and Adams, Nikolaus and Sagaut, Pierre},
  year={2009},
  publisher={Springer Science \& Business Media}
}

@article{bodart2011wall,
  title={Wall-modeled large eddy simulation in complex geometries with application to high-lift devices},
  author={Bodart, J and Larsson, J},
  journal={Annual Research Briefs},
  pages={37--48},
  year={2011},
  publisher={Center for Turbulence Research, Stanford University Stanford, CA}
}

@article{larsson2016large,
  title={Large eddy simulation with modeled wall-stress: recent progress and future directions},
  author={Larsson, Johan and Kawai, Soshi and Bodart, Julien and Bermejo-Moreno, Ivan},
  journal={Mechanical Engineering Reviews},
  volume={3},
  number={1},
  pages={15--00418},
  year={2016},
  publisher={The Japan Society of Mechanical Engineers}
}

@article{duraisamy2019turbulence,
  title={Turbulence modeling in the age of data},
  author={Duraisamy, Karthik and Iaccarino, Gianluca and Xiao, Heng},
  journal={Annual review of fluid mechanics},
  volume={51},
  number={1},
  pages={357--377},
  year={2019},
  publisher={Annual Reviews}
}

@article{spalart2000strategies,
  title={Strategies for turbulence modelling and simulations},
  author={Spalart, Philippe R},
  journal={International journal of heat and fluid flow},
  volume={21},
  number={3},
  pages={252--263},
  year={2000},
  publisher={Elsevier}
}

@article{slotnick18332cfd,
  title={CFD vision 2030 study: A path to revolutionary computational aerosciences (2014)},
  author={Slotnick, Jeffrey and Khodadoust, Abdollah and Alonso, Juan and Darmofal, David and Gropp, William and Lurie, Elizabeth and Mavriplis, DJ},
  journal={No. NF1676L-18332. Available online: https://ntrs. nasa. gov/citations/20140003093 (accessed on 26 January 2025)}
}

@article{Fedorov2011, 
author={Fedorov, A.}, 
title={Transition and stability of high-speed boundary layers}, 
journal={Annu. Rev. Fluid Mech.}, 
volume={43}, 
pages={79--95}, 
year={2011}
}

@misc{peter2001computational,
  title={Computational Aerothermodynamics in Aeroassist Applications},
  author={Peter A, Gnoffo},
  year={2001},
  publisher={NASA Langley Technical Report Server}
}

@article{gaitonde2015progress,
  title={Progress in shock wave/boundary layer interactions},
  author={Gaitonde, Datta V},
  journal={Progress in Aerospace Sciences},
  volume={72},
  pages={80--99},
  year={2015},
  publisher={Elsevier}
}

@article{dolling2001fifty,
  title={Fifty years of shock-wave/boundary-layer interaction research: what next?},
  author={Dolling, David S},
  journal={AIAA journal},
  volume={39},
  number={8},
  pages={1517--1531},
  year={2001}
}

@article{jiang1996efficient,
  title={Efficient implementation of weighted ENO schemes},
  author={Jiang, Guang-Shan and Shu, Chi-Wang},
  journal={Journal of computational physics},
  volume={126},
  number={1},
  pages={202--228},
  year={1996},
  publisher={Elsevier}
}

@article{pirozzoli2011numerical,
  title={Numerical methods for high-speed flows},
  author={Pirozzoli, Sergio},
  journal={Annual review of fluid mechanics},
  volume={43},
  number={1},
  pages={163--194},
  year={2011},
  publisher={Annual Reviews}
}

@article{zhang2024efficient,
  title={Efficient active flow control strategy for confined square cylinder wake using deep learning-based surrogate model and reinforcement learning},
  author={Zhang, Meng and Yousif, Mustafa Z and Xu, Minze and Zhou, Haifeng and Yu, Linqi and Lim, HeeChang},
  journal={arXiv preprint arXiv:2408.14232},
  year={2024}
}

@article{weiner2025model,
  title={Model-based deep reinforcement learning for accelerated learning from flow simulations},
  author={Weiner, Andre and Geise, Janis},
  journal={Meccanica},
  volume={60},
  number={6},
  pages={1771--1788},
  year={2025},
  publisher={Springer}
}

@article{schulman2017ppo,
  title={Proximal Policy Optimization Algorithms},
  author={Schulman, J. and others},
  journal={arXiv preprint arXiv:1707.06347},
  year={2017}
}

@article{lillicrap2016ddpg,
  title={Continuous control with deep reinforcement learning},
  author={Lillicrap, T. P. and others},
  journal={ICLR},
  year={2016}
}

@article{haarnoja2018sac,
  title={Soft Actor-Critic: Off-Policy Maximum Entropy Deep Reinforcement Learning},
  author={Haarnoja, T. and others},
  journal={ICML},
  year={2018}
}

@article{poulinakis2023deep,
  title={Deep learning reconstruction of pressure fluctuations in supersonic shock--boundary layer interaction},
  author={Poulinakis, Konstantinos and Drikakis, Dimitris and Kokkinakis, Ioannis William and Spottswood, S Michael},
  journal={Physics of Fluids},
  volume={35},
  number={7},
  year={2023},
  publisher={AIP Publishing}
}

@article{ren2024adaptive,
  title={Adaptive control of transonic buffet and buffeting flow with deep reinforcement learning},
  author={Ren, Kai and Gao, Chuanqiang and Xiong, Neng and Zhang, Weiwei},
  journal={Physics of Fluids},
  volume={36},
  number={1},
  year={2024},
  publisher={AIP Publishing}
}

@article{menon2026scientific,
  title={On Scientific Foundation Models: Rigorous Definitions, Key Applications, and a Comprehensive Survey},
  author={Menon, Sidharth S and Mondal, Trishit and Brahmachary, Shuvayan and Panda, Aniruddha and Joshi, Subodh M and Kalyanaraman, Kaushic and Jagtap, Ameya D},
  journal={Neural Networks},
  pages={108567},
  volume={198},
  year={2026},
  publisher={Elsevier}
}

@article{gong2024active,
  title={Active control of transonic airfoil flutter using synthetic jets through deep reinforcement learning},
  author={Gong, Tianchi and Wang, Yan and Zhao, Xiang},
  journal={Physics of Fluids},
  volume={36},
  number={10},
  year={2024},
  publisher={AIP Publishing}
}

@article{deng2023closed,
  title={Closed-loop control of transonic buffet using active shock control bump},
  author={Deng, Feng and Zhang, Shenghua and Qin, Ning},
  journal={Aerospace},
  volume={10},
  number={6},
  pages={537},
  year={2023},
  publisher={MDPI}
}

@article{gao2025transonic,
  title={Transonic buffet flow adaptive control with time-variant reduced-order model},
  author={Gao, Chuanqiang and Yang, Xinyu and Ren, Kai and Zhang, Weiwei},
  journal={Journal of Fluid Mechanics},
  volume={1022},
  pages={A32},
  year={2025},
  publisher={Cambridge University Press}
}

@article{min2024machine,
  title={Machine Learning-Based Backpressure Unstart Prediction and Warning Method for Combined Cycle Engine Hypersonic Inlet-Oriented Wide Speed Range},
  author={Min, Ke and Hong, Tanbao and Cai, Zejun and Yu, Lianchen and Zhu, Chengxiang and Zeng, Jianping},
  journal={International Journal of Aerospace Engineering},
  volume={2024},
  number={1},
  pages={2284914},
  year={2024},
  publisher={Wiley Online Library}
}

@article{sethuraman2021control,
  title={Control of the oscillations of shock train using boundary layer suction},
  author={Sethuraman, Vignesh Ram Petha and Kim, Tae Ho and Kim, Heuy Dong},
  journal={Aerospace Science and Technology},
  volume={118},
  pages={107012},
  year={2021},
  publisher={Elsevier}
}

@article{hahn2025control,
  title={Control of Shock Train Structure in Isolator by Electrical Discharge},
  author={Hahn, Loren C and Lax, Philip A and Morris, Scott C and Leonov, Sergey B and Ombrello, Timothy M and Braun, Erik L and McGann, Brendan J},
  journal={AIAA Journal},
  pages={1--13},
  year={2025},
  publisher={American Institute of Aeronautics and Astronautics}
}

@article{liang2025research,
  title={Research on Optimal Control Strategy and Method for Scramjet Fuel Pulse Injection},
  author={Liang, Shuang and Tian, Ye and Yang, Maotao and Yan, Ming and Song, Wenyan and Le, Jialing},
  journal={AIAA Journal},
  pages={1--20},
  year={2025},
  publisher={American Institute of Aeronautics and Astronautics}
}

@article{holman2005formation,
  title={Formation criterion for synthetic jets},
  author={Holman, Ryan and Utturkar, Yogen and Mittal, Rajat and Smith, Barton L and Cattafesta, Louis},
  journal={AIAA journal},
  volume={43},
  number={10},
  pages={2110--2116},
  year={2005}
}

@article{shimomura2020closed,
  title={Closed-loop flow separation control using the deep Q network over airfoil},
  author={Shimomura, Satoshi and Sekimoto, Satoshi and Oyama, Akira and Fujii, Kozo and Nishida, Hiroyuki},
  journal={AIAA Journal},
  volume={58},
  number={10},
  pages={4260--4270},
  year={2020},
  publisher={American Institute of Aeronautics and Astronautics}
}

@article{paris2023reinforcement,
  title={Reinforcement-learning-based actuator selection method for active flow control},
  author={Paris, Romain and Beneddine, Samir and Dandois, Julien},
  journal={Journal of Fluid Mechanics},
  volume={955},
  pages={A8},
  year={2023},
  publisher={Cambridge University Press}
}

@article{suarez2025flow,
  title={Flow control of three-dimensional cylinders transitioning to turbulence via multi-agent reinforcement learning},
  author={Su{\'a}rez, Pol and Alc{\'a}ntara-{\'A}vila, Francisco and Rabault, Jean and Mir{\'o}, Arnau and Font, Bernat and Lehmkuhl, Oriol and Vinuesa, Ricardo},
  journal={Communications Engineering},
  volume={4},
  number={1},
  pages={113},
  year={2025},
  publisher={Nature Publishing Group UK London}
}

@article{he2023policy,
  title={Policy transfer of reinforcement learning-based flow control: From two-to three-dimensional environment},
  author={He, Xian-Jun and Wang, Yi-Zhe and Hua, Yue and Chen, Zhi-Hua and Li, Yu-Bai and Wu, Wei-Tao},
  journal={Physics of Fluids},
  volume={35},
  number={5},
  year={2023},
  publisher={AIP Publishing}
}

@article{menon2026fekan,
  title={FEKAN: Feature-Enriched Kolmogorov--Arnold Networks},
  author={Menon, Sidharth S and Jagtap, Ameya D},
  journal={Computer Methods in Applied Mechanics and Engineering},
  volume={461},
  pages={119141},
  year={2026},
  publisher={Elsevier}
}

@article{brunton2016discovering,
  title={Discovering governing equations from data by sparse identification of nonlinear dynamical systems},
  author={Brunton, Steven L and Proctor, Joshua L and Kutz, J Nathan},
  journal={Proceedings of the national academy of sciences},
  volume={113},
  number={15},
  pages={3932--3937},
  year={2016},
  publisher={National Academy of Sciences}
}

@inproceedings{liu2025kan,
  title={KAN: Kolmogorov--arnold networks},
  author={Liu, Ziming and Wang, Yixuan and Vaidya, Sachin and Ruehle, Fabian and Halverson, James and Soljacic, Marin and Hou, Thomas and Tegmark, Max},
  booktitle={International conference on learning representations},
  volume={2025},
  pages={70367--70413},
  year={2025}
}

@inproceedings{verma2018programmatically,
  title={Programmatically interpretable reinforcement learning},
  author={Verma, Abhinav and Murali, Vijayaraghavan and Singh, Rishabh and Kohli, Pushmeet and Chaudhuri, Swarat},
  booktitle={International conference on machine learning},
  pages={5045--5054},
  year={2018},
  organization={PMLR}
}

@article{bastani2018verifiable,
  title={Verifiable reinforcement learning via policy extraction},
  author={Bastani, Osbert and Pu, Yewen and Solar-Lezama, Armando},
  journal={Advances in neural information processing systems},
  volume={31},
  year={2018}
}

@article{mnih2015human,
  title={Human-level control through deep reinforcement learning},
  author={Mnih, Volodymyr and Kavukcuoglu, Koray and Silver, David and Rusu, Andrei A and Veness, Joel and Bellemare, Marc G and Graves, Alex and Riedmiller, Martin and Fidjeland, Andreas K and Ostrovski, Georg and others},
  journal={nature},
  volume={518},
  number={7540},
  pages={529--533},
  year={2015},
  publisher={Nature Publishing Group UK London}
}

@article{guastoni2023deep,
  title={Deep reinforcement learning for turbulent drag reduction in channel flows},
  author={Guastoni, Luca and Rabault, Jean and Schlatter, Philipp and Azizpour, Hossein and Vinuesa, Ricardo},
  journal={The European Physical Journal E},
  volume={46},
  number={4},
  pages={27},
  year={2023},
  publisher={Springer}
}

@article{wang2024dynamic,
  title={Dynamic feature-based deep reinforcement learning for flow control of circular cylinder with sparse surface pressure sensing},
  author={Wang, Qiulei and Yan, Lei and Hu, Gang and Chen, Wenli and Rabault, Jean and Noack, Bernd R},
  journal={Journal of Fluid Mechanics},
  volume={988},
  pages={A4},
  year={2024},
  publisher={Cambridge University Press}
}

@inproceedings{fujimoto2018addressing,
  title={Addressing function approximation error in actor-critic methods},
  author={Fujimoto, Scott and Hoof, Herke and Meger, David},
  booktitle={International conference on machine learning},
  pages={1587--1596},
  year={2018},
  organization={PMLR}
}

@article{mondal2025shocks,
  title={Shocks Under Control: Taming Transonic Compressible Flow over an RAE2822 Airfoil with Deep Reinforcement Learning},
  author={Mondal, Trishit and Vinuesa, Ricardo and Jagtap, Ameya D},
  journal={arXiv preprint arXiv:2511.07564},
  year={2025}
}

@book{kutz2022drlfluids,
  title={Machine Learning Control of Fluid Dynamics},
  author={Kutz, J. N.},
  publisher={Cambridge University Press},
  year={2022}
}

@article{ye2024reducing,
  title={Reducing flow fluctuation using deep reinforcement learning with a CNN-based flow feature model},
  author={Ye, Shuran and Zhang, Zhen and Wang, Yiwei and Huang, Chenguang},
  journal={Ocean Engineering},
  volume={306},
  pages={118089},
  year={2024},
  publisher={Elsevier}
}

@article{dong2023surrogate,
  title={Surrogate model-based deep reinforcement learning for experimental study of active flow control of circular cylinder},
  author={Dong, Xinhui and Hong, Hao and Deng, Xiaolong and Zhong, Wei and Hu, Gang},
  journal={Physics of Fluids},
  volume={35},
  number={10},
  year={2023},
  publisher={AIP Publishing}
}

@article{mao2024model,
  title={Model-based deep reinforcement learning for active control of flow around a circular cylinder using action-informed episode-based neural ordinary differential equations},
  author={Mao, Yiqian and Zhong, Shan and Yin, Hujun},
  journal={Physics of Fluids},
  volume={36},
  number={8},
  year={2024},
  publisher={AIP Publishing}
}

@inproceedings{werner2024numerical,
  title={Numerical evidence for sample efficiency of model-based over model-free reinforcement learning control of partial differential equations},
  author={Werner, Stefan and Peitz, Sebastian},
  booktitle={2024 European Control Conference (ECC)},
  pages={2965--2971},
  year={2024},
  organization={IEEE}
}

@article{janner2019trust,
  title={When to trust your model: Model-based policy optimization},
  author={Janner, Michael and Fu, Justin and Zhang, Marvin and Levine, Sergey},
  journal={Advances in neural information processing systems},
  volume={32},
  year={2019}
}

@article{chua2018deep,
  title={Deep reinforcement learning in a handful of trials using probabilistic dynamics models},
  author={Chua, Kurtland and Calandra, Roberto and McAllister, Rowan and Levine, Sergey},
  journal={Advances in neural information processing systems},
  volume={31},
  year={2018}
}

@article{moerland2023model,
  title={Model-based reinforcement learning: A survey},
  author={Moerland, Thomas M and Broekens, Joost and Plaat, Aske and Jonker, Catholijn M and others},
  journal={Foundations and Trends{\textregistered} in Machine Learning},
  volume={16},
  number={1},
  pages={1--118},
  year={2023},
  publisher={Now Publishers, Inc.}
}

@article{hafner2020mastering,
  title={Mastering atari with discrete world models},
  author={Hafner, Danijar and Lillicrap, Timothy and Norouzi, Mohammad and Ba, Jimmy},
  journal={arXiv preprint arXiv:2010.02193},
  year={2020}
}

@article{arslan2025surrogate,
  title={Surrogate-Based Deep Reinforcement Learning for Active Flow Control on Slender Bodies},
  author={Arslan, K{\i}van{\c{c}} and {\"O}zgen, Serkan},
  journal={Journal of Spacecraft and Rockets},
  pages={1--13},
  year={2025},
  publisher={American Institute of Aeronautics and Astronautics}
}

@article{portal2023active,
  title={Active flow control on airfoils by reinforcement learning},
  author={Portal-Porras, Koldo and Fernandez-Gamiz, Unai and Zulueta, Ekaitz and Garcia-Fernandez, Roberto and Berrizbeitia, Saioa Etxebarria},
  journal={Ocean Engineering},
  volume={287},
  pages={115775},
  year={2023},
  publisher={Elsevier}
}

@article{li2021learning,
  title={Learning the aerodynamic design of supercritical airfoils through deep reinforcement learning},
  author={Li, Runze and Zhang, Yufei and Chen, Haixin},
  journal={AIAA Journal},
  volume={59},
  number={10},
  pages={3988--4001},
  year={2021},
  publisher={American Institute of Aeronautics and Astronautics}
}

@article{halder2020deep,
  title={Deep learning based reduced order model for airfoil-gust and aeroelastic interaction},
  author={Halder, Rahul and Damodaran, Murali and Khoo, BC},
  journal={AIAA Journal},
  volume={58},
  number={10},
  pages={4304--4321},
  year={2020},
  publisher={American Institute of Aeronautics and Astronautics}
}

@article{tao2025control,
  title={Control of Hypersonic Shock-Wave/Laminar-Boundary-Layer Interaction Using Deep Reinforcement Learning},
  author={Tao, Shancong and Shi, Xiaotian and Xiong, Xue-Lu and Zhou, Yi},
  journal={AIAA Journal},
  volume={63},
  number={10},
  pages={4308--4322},
  year={2025},
  publisher={American Institute of Aeronautics and Astronautics}
}

@article{xia2024active,
  title={Active flow control for bluff body drag reduction using reinforcement learning with partial measurements},
  author={Xia, Chengwei and Zhang, Junjie and Kerrigan, Eric C and Rigas, Georgios},
  journal={Journal of Fluid Mechanics},
  volume={981},
  pages={A17},
  year={2024},
  publisher={Cambridge University Press}
}

@article{fan2020reinforcement,
  title={Reinforcement learning for bluff body active flow control in experiments and simulations},
  author={Fan, Dixia and Yang, Liu and Wang, Zhicheng and Triantafyllou, Michael S and Karniadakis, George Em},
  journal={Proceedings of the National Academy of Sciences},
  volume={117},
  number={42},
  pages={26091--26098},
  year={2020},
  publisher={National Academy of Sciences}
}

@article{zong2025closed,
  title={Closed-loop supersonic flow control with a high-speed experimental deep reinforcement learning framework},
  author={Zong, Haohua and Wu, Yun and Li, Jinping and Su, Zhi and Liang, Hua},
  journal={Journal of Fluid Mechanics},
  volume={1009},
  pages={A3},
  year={2025},
  publisher={Cambridge University Press}
}

@inproceedings{Tobin2017,
  author    = {Tobin, J. and Fong, R. and Ray, A. and Schneider, J. and Zaremba, W. and Abbeel, P.},
  title     = {Domain randomization for transferring deep neural networks from simulation to the real world},
  booktitle = {IEEE/RSJ International Conference on Intelligent Robots and Systems (IROS)},
  year      = {2017}
}

@article{Paris2021,
  author  = {Paris, R. and Beneddine, S. and Dandois, J.},
  title   = {Robust flow control and optimal sensor placement using deep reinforcement learning},
  journal = {Journal of Fluid Mechanics},
  volume  = {913},
  pages   = {A15},
  year    = {2021}
}

@inproceedings{Finn2017,
  author    = {Finn, C. and Abbeel, P. and Levine, S.},
  title     = {Model-agnostic meta-learning for fast adaptation of deep networks},
  booktitle = {Proceedings of the 34th International Conference on Machine Learning},
  year      = {2017}
}

@article{kochkov2021machine,
  title={Machine learning--accelerated computational fluid dynamics},
  author={Kochkov, Dmitrii and Smith, Jamie A and Alieva, Ayya and Wang, Qing and Brenner, Michael P and Hoyer, Stephan},
  journal={Proceedings of the National Academy of Sciences},
  volume={118},
  number={21},
  pages={e2101784118},
  year={2021},
  publisher={National Academy of Sciences}
}

@article{novati2021automating,
  title={Automating turbulence modelling by multi-agent reinforcement learning},
  author={Novati, Guido and de Laroussilhe, Hugues Lascombes and Koumoutsakos, Petros},
  journal={Nature Machine Intelligence},
  volume={3},
  number={1},
  pages={87--96},
  year={2021},
  publisher={Nature Publishing Group UK London}
}

@article{gautier2015closed,
  title={Closed-loop separation control using machine learning},
  author={Gautier, Nicolas and Aider, J-L and Duriez, THOMAS and Noack, Bend R and Segond, Marc and Abel, Markus},
  journal={Journal of Fluid Mechanics},
  volume={770},
  pages={442--457},
  year={2015},
  publisher={Cambridge University Press}
}

@article{wan2021rlcontrol,
  title={Reinforcement learning for flow control: advances and open challenges},
  author={Wan, Z. and others},
  journal={Annual Review of Fluid Mechanics},
  volume={53},
  pages={305--334},
  year={2021}
}

@inproceedings{greydanus2018visualizing,
  title={Visualizing and understanding Atari agents},
  author={Greydanus, S. and others},
  booktitle={ICML},
  year={2018}
}

@article{hinton2008visualizing,
  title={Visualizing data using t-SNE},
  author={van der Maaten, L. and Hinton, G.},
  journal={Journal of Machine Learning Research},
  volume={9},
  pages={2579–2605},
  year={2008}
}

@book{pearl2009causality,
  title={Causality: Models, Reasoning and Inference},
  author={Pearl, J.},
  publisher={Cambridge University Press},
  year={2009}
}

@article{raissi2019physics,
  title={Physics-informed neural networks: A deep learning framework for solving forward and inverse problems involving nonlinear partial differential equations},
  author={Raissi, Maziar and Perdikaris, Paris and Karniadakis, George E},
  journal={Journal of Computational physics},
  volume={378},
  pages={686--707},
  year={2019},
  publisher={Elsevier}
}

@article{brunton2020data,
  title={Data-driven science and engineering: Machine learning, dynamical systems, and control},
  author={Brunton, S. L. and Kutz, J. N.},
  journal={Cambridge University Press},
  year={2020}
}

@book{pope2000turbulent,
  title={Turbulent Flows},
  author={Pope, S. B.},
  publisher={Cambridge University Press},
  year={2000}
}

@article{bewley2018feedback,
  title={Feedback control: A perspective for the fluid dynamicist},
  author={Bewley, T. and others},
  journal={Annual Review of Fluid Mechanics},
  volume={50},
  pages={357--385},
  year={2018}
}

@article{banerjee2025survey,
  title={A survey on physics informed reinforcement learning: Review and open problems},
  author={Banerjee, Chayan and Nguyen, Kien and Fookes, Clinton and Raissi, Maziar},
  journal={Expert Systems with Applications},
  pages={128166},
  year={2025},
  publisher={Elsevier}
}

@book{rudin1991inverse,
  title={Inverse Problems in Imaging},
  author={Rudin, L. I. and others},
  publisher={Springer},
  year={1991}
}

@book{sutton2018reinforcement,
  title={Reinforcement Learning: An Introduction},
  author={Sutton, R. S. and Barto, A. G.},
  publisher={MIT Press},
  year={2018}
}

@inproceedings{furst2019event,
  title={Event-triggered reinforcement learning},
  author={F{\"u}rst, J. and others},
  booktitle={IEEE Conference on Decision and Control},
  year={2019}
}

@article{liu2025model,
  title={Model-based reinforcement learning for control of strongly disturbed unsteady aerodynamic flows},
  author={Liu, Zhecheng and Beckers, Diederik and Eldredge, Jeff D},
  journal={AIAA Journal},
  pages={1--21},
  year={2025},
  publisher={American Institute of Aeronautics and Astronautics}
}

@article{van2016deep,
  title={Deep reinforcement learning for robotic manipulation with asynchronous policy updates},
  author={Lillicrap, T. P. and others},
  journal={ICLR},
  year={2016}
}

@article{ng1999policy,
  title={Policy invariance under reward transformations: Theory and application to reward shaping},
  author={Ng, A. Y. and Russell, S. J.},
  journal={ICML},
  year={1999}
}

@article{brunton2015closed,
  title={Closed-loop turbulence control: Progress and challenges},
  author={Brunton, Steven L and Noack, Bernd R},
  journal={Applied Mechanics Reviews},
  volume={67},
  number={5},
  pages={050801},
  year={2015},
  publisher={American Society of Mechanical Engineers}
}

@inproceedings{wilson2007multi,
  title={Multi-task reinforcement learning: a hierarchical bayesian approach},
  author={Wilson, Aaron and Fern, Alan and Ray, Soumya and Tadepalli, Prasad},
  booktitle={Proceedings of the 24th international conference on Machine learning},
  pages={1015--1022},
  year={2007}
}

@inproceedings{sodhani2021multi,
  title={Multi-task reinforcement learning with context-based representations},
  author={Sodhani, Shagun and Zhang, Amy and Pineau, Joelle},
  booktitle={International conference on machine learning},
  pages={9767--9779},
  year={2021},
  organization={PMLR}
}

@article{yan2025deep,
  title={Deep reinforcement cross-domain transfer learning of active flow control for three-dimensional bluff body flow},
  author={Yan, Lei and Wang, Qiulei and Hu, Gang and Chen, Wenli and Noack, Bernd R},
  journal={Journal of Computational Physics},
  volume={529},
  pages={113893},
  year={2025},
  publisher={Elsevier}
}

@article{silver2021reward,
  title={Reward is enough},
  author={Silver, David and Singh, Satinder and Precup, Doina and Sutton, Richard S},
  journal={Artificial intelligence},
  volume={299},
  pages={103535},
  year={2021},
  publisher={Elsevier}
}

\end{document}